%% file: main.tex
\documentclass[10pt,twocolumn,letterpaper]{article}

\usepackage[pagenumbers]{cvpr}

\usepackage[accsupp]{axessibility}  

\newcommand\numberthis{\addtocounter{equation}{1}\tag{\theequation}}

\usepackage{textcomp, gensymb}
\usepackage{times}

\usepackage{epsfig}
\usepackage{graphicx}
\usepackage[belowskip=-10pt,aboveskip=2pt,font=small]{caption}
\usepackage{dblfloatfix}

\usepackage{amsmath}
\usepackage{amssymb}
\usepackage{nccmath}
\usepackage{bm}

\usepackage{multirow}
\usepackage{rotating}
\usepackage{booktabs}
\usepackage{algorithm}

\usepackage{enumerate}
\usepackage[olditem,oldenum]{paralist}

\usepackage{xcolor}
\usepackage{soul}

\usepackage{cite}

\usepackage[pagebackref,breaklinks,colorlinks]{hyperref}

\usepackage[capitalize]{cleveref}
\crefname{section}{Sec.}{Secs.}
\Crefname{section}{Section}{Sections}
\Crefname{table}{Table}{Tables}
\crefname{table}{Tab.}{Tabs.}

\def\cvprPaperID{3413}
\def\confName{CVPR}
\def\confYear{2023}

\input{preamble}

\begin{document}

\input{0_metadata}

\twocolumn[{
\renewcommand\twocolumn[1][]{#1}
\maketitle
\vspace{-0.3cm}
\input{teaser}
}]

\input{1_abstract}
\input{2_intro}
\input{3_related_work}
\input{4_dataset}
\input{5_approach}
\input{6_experiments}
\input{7_results}
\input{8_conclusion}

{\small
\bibliographystyle{unsrt}
\bibliography{macros,main}
}

\newpage
\input{X_supplementary}

\end{document}

%% file: preamble.tex
\usepackage{overpic}
\usepackage{enumitem} 
\usepackage{paralist} 
\usepackage{overpic} 
\usepackage{color}
\usepackage{tabularx}

\usepackage{listings}
\definecolor{turquoise}{cmyk}{0.65,0,0.1,0.3}
\definecolor{purple}{rgb}{0.65,0,0.65}
\definecolor{dark_green}{rgb}{0, 0.5, 0}
\definecolor{dullgreen}{rgb}{0.3, 0.8, 0.3}
\definecolor{orange}{rgb}{0.8, 0.6, 0.2}
\definecolor{red}{rgb}{0.8, 0.2, 0.2}
\definecolor{darkred}{rgb}{0.6, 0.1, 0.05}
\definecolor{blueish}{rgb}{0.3, 0.3, .6}
\definecolor{light_gray}{rgb}{0.7, 0.7, .7}
\definecolor{pink}{rgb}{1, 0, 1}
\definecolor{greyblue}{rgb}{0.25, 0.25, 1}
\definecolor{awesome}{rgb}{1.0, 0.13, 0.32}
\definecolor{flexyellow}{rgb}{0.894, 0.663, 0.204}
\definecolor{goalblue}{rgb}{0.063, 0.207, 0.894}
\definecolor{sagaorange}{rgb}{0.894, 0.365, 0.086}

\definecolor{myyellow}{rgb}{0.71, 0.67, 0.51}
\definecolor{myblue}{rgb}{0.57, 0.63, 0.72}
\definecolor{mypink}{rgb}{0.67, 0.55, 0.54}
\newcommand{\dullgreen}[1]{\textcolor{dullgreen}{#1}}
\newcommand{\red}[1]{\textcolor{red}{#1}}

\newcommand{\goalblue}[1]{\textcolor{goalblue}{#1}}
\newcommand{\flexyellow}[1]{\textcolor{flexyellow}{#1}}
\newcommand{\sagaorange}[1]{\textcolor{sagaorange}{#1}}

\newcommand{\flex}[0]{{\fontfamily{qcr}\selectfont FLEX}}

\usepackage{blindtext}

\renewcommand{\paragraph}[1]{\vspace{1em}\noindent\textbf{#1}.}

\usepackage{gensymb}

\usepackage{graphicx}               
\usepackage{sidecap}

\everypar{\looseness=-1}
\usepackage{dsfont}

%% file: 0_metadata.tex
\title{\flex: Full-Body Grasping Without Full-Body Grasps}

\author{Purva Tendulkar\hspace{.2cm}
D\'idac Sur\'is\hspace{.2cm}
Carl Vondrick\\[0.09cm]
Columbia University\\[0.01cm]
{\tt\footnotesize \{purvaten, didacsuris, vondrick\}@cs.columbia.edu}
}

%% file: teaser.tex
\begin{center}
\vspace{-0.5cm}
\centering
\captionsetup{type=figure}
\includegraphics[width=1\linewidth]{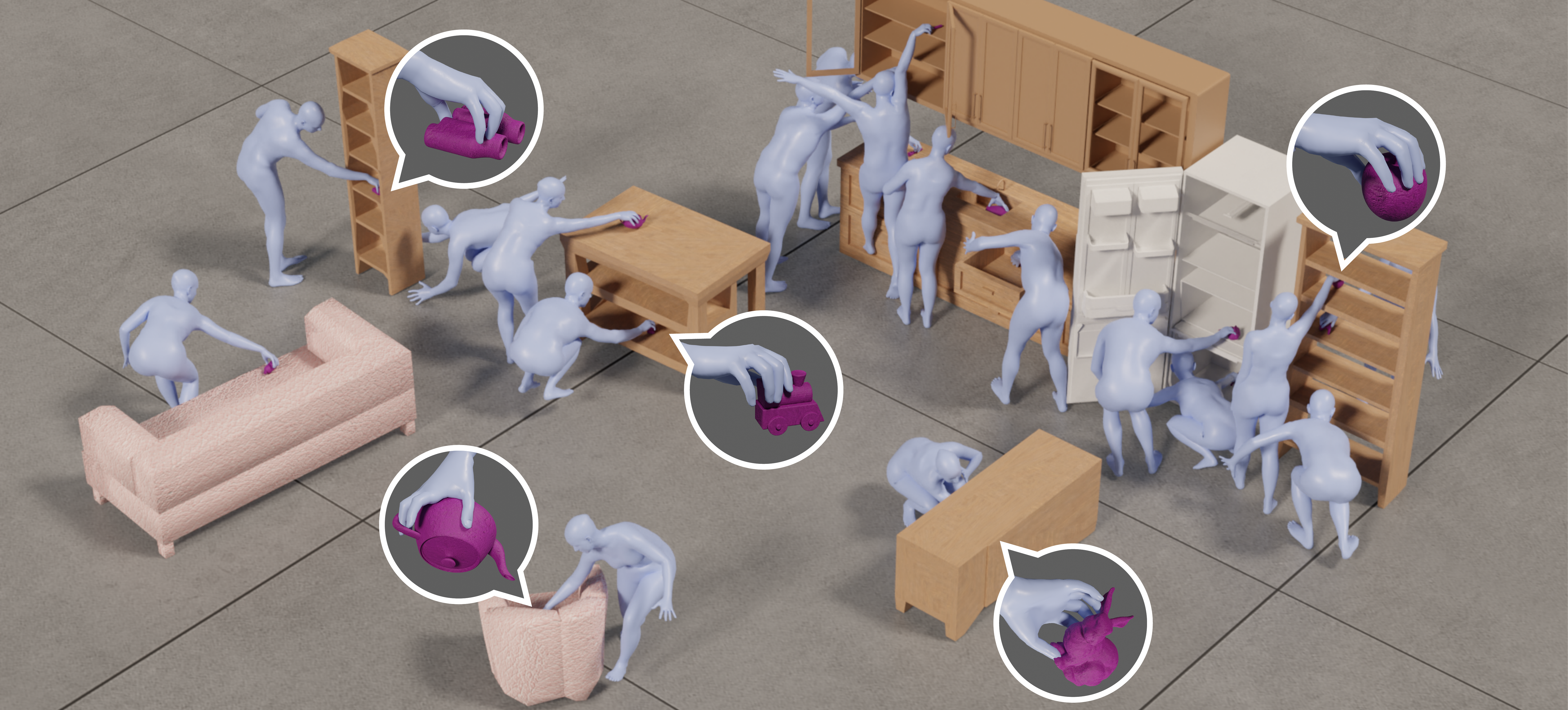}
\captionof{figure}{\flex~generates diverse full-body poses for grasping 3D objects in complex scenes. It does not use any full-body grasping data, and fully relies on body-pose priors (without grasps) and hand grasping priors (without full-body poses).}

\vspace{0.3cm}

\label{fig:teaser}
\end{center}

%% file: 1_abstract.tex
\begin{abstract}
    \vspace{-0.5cm}
   Synthesizing 3D human avatars interacting realistically with a scene is an important problem with applications in AR/VR, video games, and robotics. Towards this goal, we address the task of generating a virtual human -- hands and full body -- grasping everyday objects. Existing methods approach this problem by collecting a 3D dataset of humans interacting with objects and training on this data. However, 1) these methods do not generalize to different object positions and orientations or to the presence of furniture in the scene, and 2) the diversity of their generated full-body poses is very limited. In this work, we address all the above challenges to generate realistic, diverse full-body grasps in everyday scenes without requiring any 3D full-body grasping data. Our key insight is to leverage the existence of both full-body pose and hand-grasping priors, composing them using 3D geometrical constraints to obtain full-body grasps. We empirically validate that these constraints can generate a variety of feasible human grasps that are superior to baselines both quantitatively and qualitatively.
   See our webpage for more details: \href{https://flex.cs.columbia.edu/}{flex.cs.columbia.edu}.
  \vspace{-0.3cm}
\end{abstract}

%% file: 2_intro.tex
\vspace{-1.2cm}
\section{Introduction}
\label{sec:intro}
\vspace{-0.16cm}
Generating realistic virtual humans is an exciting step towards building better animation tools, video games, immersive VR technology and more realistic simulators with human presence.
Towards this goal, the research community has invested a lot of effort in collecting large-scale 3D datasets of humans~\cite{AMASS:2019, ACCAD, DanceDB:Aristidou:2019, BMLhandball, BMLrub, cmuWEB, dfaust:CVPR:2017, Eyes_Japan, ghorbani2020movi, chatzitofis2020human4d, HEva_Sigal:IJCV:10b, KIT_Dataset, MoSh_lopermahmoodetal2014, MPI_HDM05, PosePrior_Akhter:CVPR:2015, SFU, TCD_hands, TotalCapture_Trumble:BMVC:2017}.
However, the reliance on data collection will be a major bottleneck when scaling to broader scenarios, for two main reasons. 
First, data collection using optical marker-based motion capture (MoCap) systems is quite tedious to work with. This becomes even more complicated when objects~\cite{GRAB:2020} or scenes~\cite{PROX:2019} are involved, requiring expertise in specialized hardware systems \cite{structure,kinect4xbox,infra_vantage}, as well as commercial software~\cite{vicon, poser, skanect, monocle}.
Even with the best combination of state-of-the-art solutions, this process often requires multiple skilled technicians to ensure clean data~\cite{GRAB:2020}.

Second, it is practically impossible to capture all possible ways of interacting with the ever-changing physical world. The number of scenarios grows exponentially with every considered variable (such as human pose, object class, task, or scene characteristics). For this reason, models trained on task-specific datasets suffer from the limitations of the data. For example, methods that are supervised on the GRAB dataset~\cite{GRAB:2020} for full-body grasping~\cite{taheri2021goal, wu2022saga} fail to grasp objects when the object position and/or orientation is changed, and generate poses with virtually no diversity.
This is understandable since the GRAB dataset mostly consists of \emph{standing} humans grasping objects at a \emph{fixed height}, interacting with them in a relatively \emph{small range} of physical motions. 
However in realistic scenarios, we expect to see objects in all sorts of configurations - lying on the floor, on the top shelf of a cupboard, inside a kitchen sink, etc.
 
To build human models that work in realistic scene configurations, we need to fundamentally re-think how to solve 3D tasks without needing any additional data, effectively utilizing existing data. 
In this paper, we address the task of generating full-body grasps for everyday objects in realistic household environments, by leveraging the success of hand grasping models~\cite{jiang2021graspTTA, GRAB:2020, turpin2022grasp, zhou2022toch} and recent advances in human body pose modeling~\cite{AMASS:2019,SMPL-X:2019}.

Our key observation is that we can compose different 3D generative models via geometrical and anatomical constraints. Having a strong prior over full-body human poses (knowing what poses are feasible and natural), when combined with strong grasping priors, allows us to express full-body \emph{grasping} poses. This combination leads to full-body poses which satisfy both priors resulting in natural poses that are suited for grasping objects, as well as hand grasps that human poses can easily match.

Our contributions are as follows. 
First, we propose \flex, a framework to generate full-body grasps without full-body grasping data. Given a pre-trained hand-only grasping model as well as a pre-trained body pose prior, we search in the latent spaces of these models to generate a human mesh whose hand matches that of the hand grasp, while simultaneously handling the constraints imposed by the scene, such as avoiding obstacles. To achieve this, we introduce a novel obstacle-avoidance loss that treats the human as a connected graph which breaks when intersected by an obstacle.
In addition, we show both quantitatively and qualitatively that \flex~allows us to generate a wide range of natural grasping poses for a variety of scenarios, greatly improving on previous approaches.
Finally, we introduce the ReplicaGrasp dataset, built by spawning 3D objects inside ReplicaCAD~\cite{szot2021habitat} scenes using Habitat \cite{habitat19iccv}.

\begin{figure}[t!]
    \centering
    \includegraphics[width=1.\linewidth]{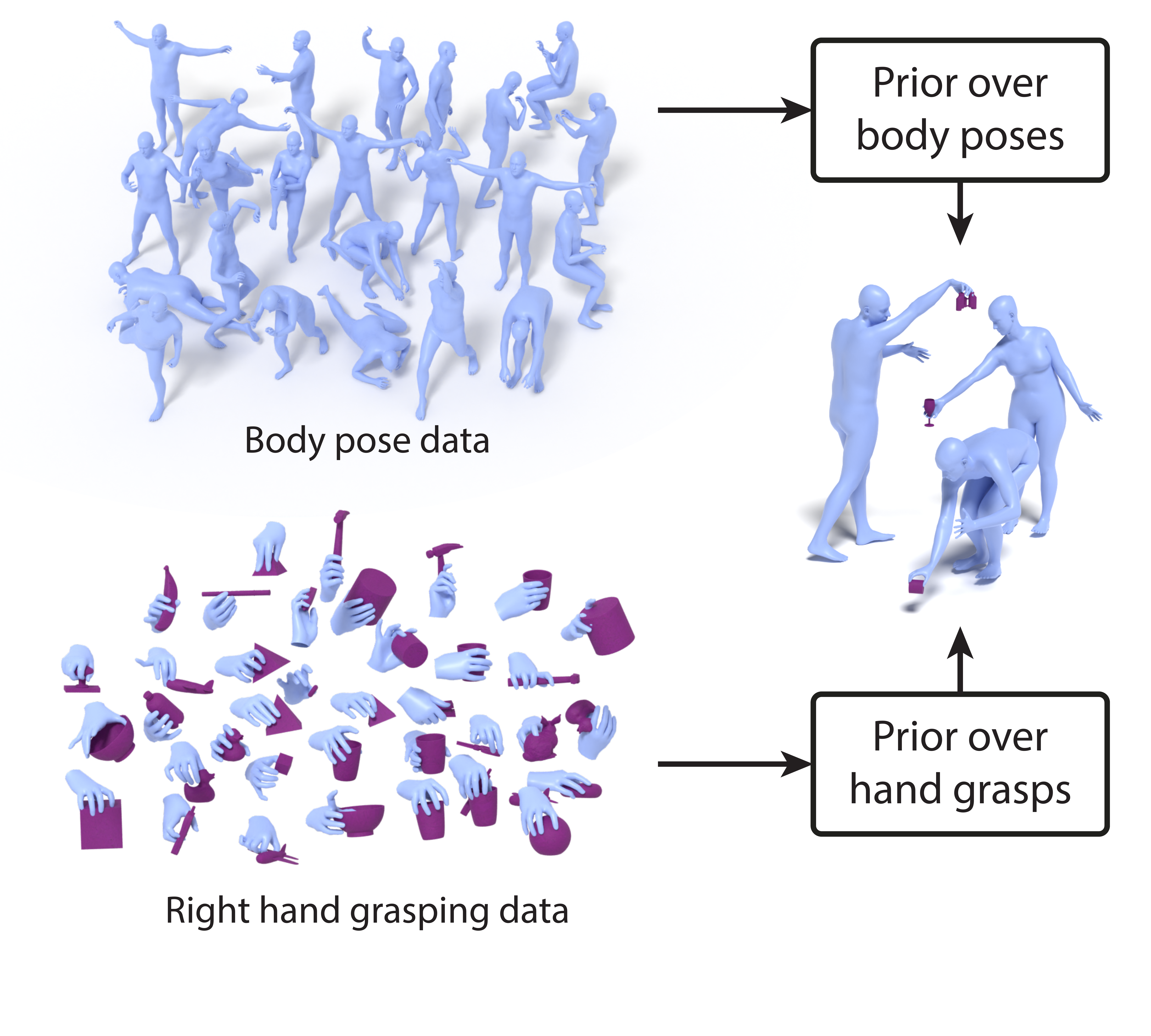}
    \vspace{-3em}
    \caption{\textbf{Overview}. We leverage existing body pose priors and hand-grasping models (left) to perform full-body grasping in complex scenes (right). Our method does not rely on any data for full-body grasping and surpasses methods requiring it, in terms of diversity and generalization to complex scenes.}
    \label{fig:intuition}
\end{figure}

%% file: 3_related_work.tex
\section{Related Work}
\label{sec:related_work}
\vspace{-0.3cm}

\begin{figure*}[t!]
    \centering
    \vspace{-0.1cm}
    \includegraphics[width=1.\linewidth]{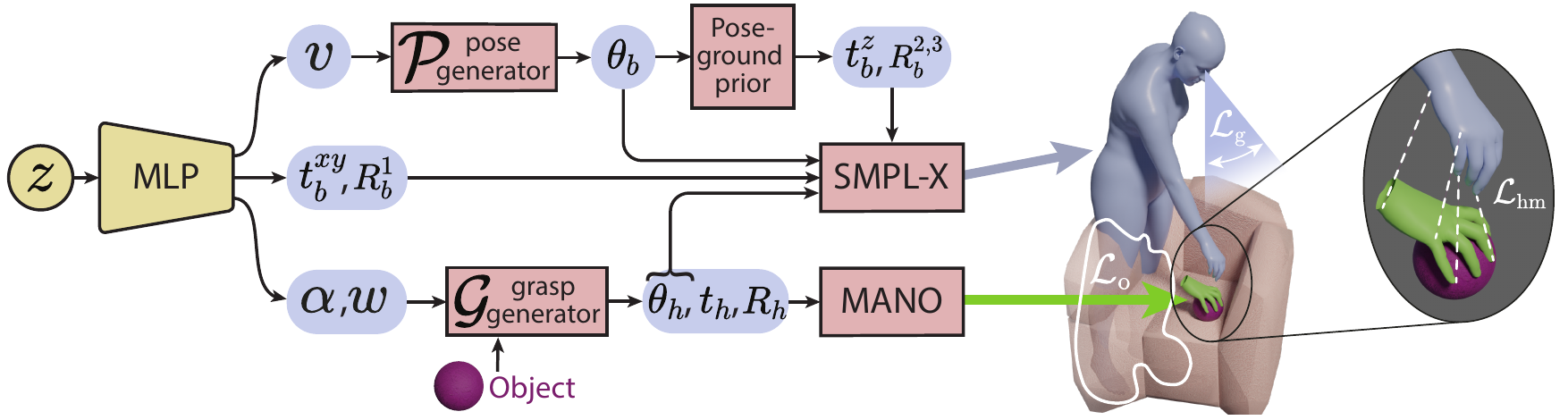}
    \caption{\textbf{Method}. Given pre-trained hand-grasping and human pose priors, our method performs a gradient-based search procedure over five different landscapes to minimize the hand matching, obstacle and gaze losses. Additionally, our data-driven pose-ground prior ensures that the pose is reasonable with respect to the ground. In the figure, the parameters we optimize are shown in \textcolor{myyellow}{yellow}, the differentiable (frozen) layers are shown in \textcolor{mypink}{pink}, and the activations are shown in \textcolor{myblue}{blue}.}
    \vspace{-0.2cm}
    \label{fig:method}
\end{figure*}

\paragraph{Object grasping} Generating a 3D hand for grasping objects is a widely studied task in robotics~\cite{CutkoskyGrasp, Detry2010RefiningGA, HsiaoImit, KrugEfficient, liu2019generating, BernardinSensor, Miller2004GraspitAV, brahmbhatt2019contactgrasp, turpin2022grasp}, graphics~\cite{Kalisiak_animate, inter_capt_synth, data_driven_grasp, PollardPhysically, RijpkemaGrasp, ManipNet, Glauser_interactive, SSundaram:2019:STAG} and 3D computer vision~\cite{jiang2021graspTTA, GRAB:2020, karunratanakul2020grasping, grady2021contactopt, brahmbhatt2020contactpose, brahmbhatt2019contactdb}.
Many works have also studied the anatomy of human hands to create grasp taxonomies~\cite{FelixGrasp, Kamakura1980PatternsOS, Napier1956ThePM, Puhlmann_grasp, SMPL-X:2019}.
Most existing work tries to imitate static hand-grasping positions from data~\cite{jiang2021graspTTA, GRAB:2020, karunratanakul2020grasping, grady2021contactopt}, while other efforts generate stable dynamic grasps using either motion-based grasping data ~\cite{zhou2022toch} or reinforcement learning \cite{christen2022d}. Recently,
Turpin \textit{et al.}~\cite{turpin2022grasp} used a differentiable physics simulator to generate grasps that are physically stable.
All these methods have looked at generating grasps for objects in presence of simple or no obstacles (\eg, placed on a counter-top). Instead, we focus on generating hand grasps for objects in more realistic scenarios (\eg., objects in refrigerator, drawers, kitchen sink).

\paragraph{Full-body grasping}
Instead of synthesizing just hand-object interactions, the community has recently started generating full-body interactions with objects. GRAB dataset~\cite{GRAB:2020} for full-body object interaction was collected using motion capture. GOAL~\cite{taheri2021goal} and SAGA~\cite{wu2022saga} use GRAB to build generative models for full-body grasping. Synthesizing full-body interactions has also been explored to create simulators (VirtualHome) for studying house-hold activities~\cite{puig2018virtualhome}. VirtualHome uses pre-defined animations but they are not very realistic and intersect with other objects. In this work, we generate realistic full-body grasps \emph{without using full-body data and in presence of obstacles.}

\paragraph{Test-time optimization}
Test-time optimization has been used to improve neural network generalization by either using task-specific constraints or self-supervision during inference~\cite{Vidit_online, shocher2018zero, sun2020test, bau2020semantic, Kalal_tracking, mullapudi2019online}.
This principle is applied to the hand-object grasping problem~\cite{jiang2021graspTTA, wu2022saga, taheri2021goal, GRAB:2020} by first generating a coarse grasp and later refining it using test-time optimization. These methods are especially suitable for tasks where constraints are easy to specify. Zhang \textit{et al.} \cite{zhang2020phosa} use ``3D common sense'' constraints to resolve ambiguity in 3D spatial arrangements of humans and objects jointly from 2D, while Liu \textit{et al.}~\cite{liu2022shadows} use shadows as a constraint for inferring the 3D structure of an occluded object. Similarly, we obtain full-body grasps by combining different priors satisfying intuitive geometrical constraints.

%% file: 4_dataset.tex
\section{ReplicaGrasp Dataset}
\label{sec:dataset}
Existing benchmarks for full-body object grasping~\cite{GRAB:2020} only consider objects in a limited height range without any other objects in their vicinity. 
To make the task more challenging and representative of the real-world, we build the ReplicaGrasp dataset. ReplicaGrasp contains 50 everyday objects from GRAB (such as wineglass or cellphone) present in 48 receptacles of ReplicaCAD~\cite{szot2021habitat}, simulated with the Habitat simulator~\cite{habitat19iccv} to be in a variety of feasible positions, leading to a total of 4.8k instances.
The receptacles include surfaces of both rigid and articulated furniture items, such as drawers, which may be open to different degrees, leading to interesting cases like objects being deep inside on the bottom shelf of refrigerator, or on the top shelf of a cupboard. 
We refer to the furniture on which the target object to-be-grasped is spawned as the `obstacle' for that instance.
Succeeding on this dataset requires generating full-body human grasps that are ``scene-aware''---not intersecting with obstacles---, as well as natural and feasible.

%% file: 5_approach.tex
\section{Approach}
\label{sec:approach}

Given a 3D object mesh and a set of 3D obstacle meshes in its vicinity, our goal is to generate a 3D human mesh grasping the object without intersecting with the obstacles.

\subsection{Preliminaries}
\vspace{-0.2cm}
\paragraph{Hand model} We use the MANO~\cite{MANO:SIGGRAPHASIA:2017} differentiable 3D hand model, that takes as input the full-finger articulated pose $\theta_{\text{h}} \in \mathbb{R}^{\mathrm{15x3}}$, the wrist translation $t_{\text{h}} \in \mathbb{R}^{\mathrm{3}}$, and the wrist global orientation $R_{\text{h}} \in \mathbb{R}^{\mathrm{3}}$, and outputs a 3D mesh $\mathcal{M}_{\text{h}}$, 
with vertices $\mathcal{V}_{\text{h}}$ in a global coordinate system.

\paragraph{Human body model}
We use the SMPL-X~\cite{SMPL-X:2019} statistical 3D whole-body model, which jointly represents the body, head, face and hands. 
SMPL-X is a differentiable function that takes as input the full-body pose $\theta_{\text{b}} \in \mathbb{R}^{\mathrm{21x3}}$, the full-finger articulated pose $\theta_{\text{h}} \in \mathbb{R}^{\mathrm{15x3}}$, the pelvis translation $t_{\text{b}} \in \mathbb{R}^{\mathrm{3}}$ and orientation $R_{\text{b}} \in \mathbb{R}^{\mathrm{3}}$, and optionally, body shape parameters
and facial expression, and outputs a 3D mesh $\mathcal{M}_{\text{b}}$, with vertices $\mathcal{V}_{\text{b}}$ in a global coordinate system. 

\vspace{-0.1cm}
\paragraph{Pre-trained generative models} We use a pre-trained generative model $\mathcal{P}$ that has learned a prior over body poses. It takes as input a latent vector $v$ and generates a body pose which can be used as input to SMPL-X. Similarly, $\mathcal{G}$ generates right hand grasps. It takes as input a latent vector $w$, an approaching angle $\alpha$ and an object $\mathcal{O}$, represented by its vertices, and generates a hand pose, including its translation and rotation, to be used as input to MANO. We implement $\mathcal{P}$ using VPoser \cite{SMPL-X:2019} and $\mathcal{G}$ using GrabNet \cite{GRAB:2020}.
\vspace{-0.04cm}

\subsection{Method}
\vspace{-0.1cm}
Given a pre-trained right hand grasping model $\mathcal{G}$ that can predict global MANO parameters $\{\theta_{\text{h}}, t_{\text{h}}, R_{\text{h}}\}$ for a given object, as well as a pre-trained pose prior $\mathcal{P}$  that can generate feasible full-body poses $\theta_{\text{b}}$, our approach called~\flex~ (\textbf{F}ull-body \textbf{L}atent \textbf{Ex}ploration)  generates a 3D human grasping the desired object. To do so,~\flex~searches in the latent spaces of $\mathcal{G}$ and $\mathcal{P}$ to find the latent variables $w$ and $v$, respectively, as well as over the space of approaching angles $\alpha$, SMPL-X translations $t_{\text{b}}$ and global orientations $R_{\text{b}}$, which are represented in `yaw-pitch-roll' format.
See~\cref{fig:method} for an overview of the method.

This search, or latent-space exploration, is done via model inversion, by backpropagating the gradient of a loss at the output of our model, and finding the inputs that minimize it. This procedure is done iteratively, until the loss is minimized. During the search, we keep the weights of $\mathcal{G}$ and $\mathcal{P}$  frozen. Therefore, we do not perform any training; the procedure described in this section takes place at inference time. We describe the losses we use next.

\subsection{Losses}

\paragraph{Hand matching loss} The main intuition of our paper is that we can combine generative models using constraints based on the geometry of their outputs. Specifically, we can combine the generated hand and body meshes because there exists a connection between them: the hands have to match.
Given vertices $\mathcal{V}_{\text{b}}$ (output from SMPL-X) and $\mathcal{V}_{\text{h}}$ (output from MANO), we align them by minimizing:
\begin{equation}
    \vspace{-0.1cm}
    \mathcal{L}_{\text{hm}} = \frac{1}{|\mathcal{V}_{\text{h}}|} \sum_{i = 1}^{|\mathcal{V}_{\text{h}}|}{d_{vv}(\mathcal{V}_{h_i}, \mathcal{V}^{h}_{b_i})},
\end{equation}
where $d_{vv}(., .)$ is the $L^2$ distance between two vertices in the 3D space. $\mathcal{V}^{h}_{b}$ represents the vertices within  $\mathcal{V}_{b}$ that correspond to the right hand---the rest are not used in this loss.

\begin{figure}[t!]
    \centering
    \includegraphics[width=1.\linewidth]{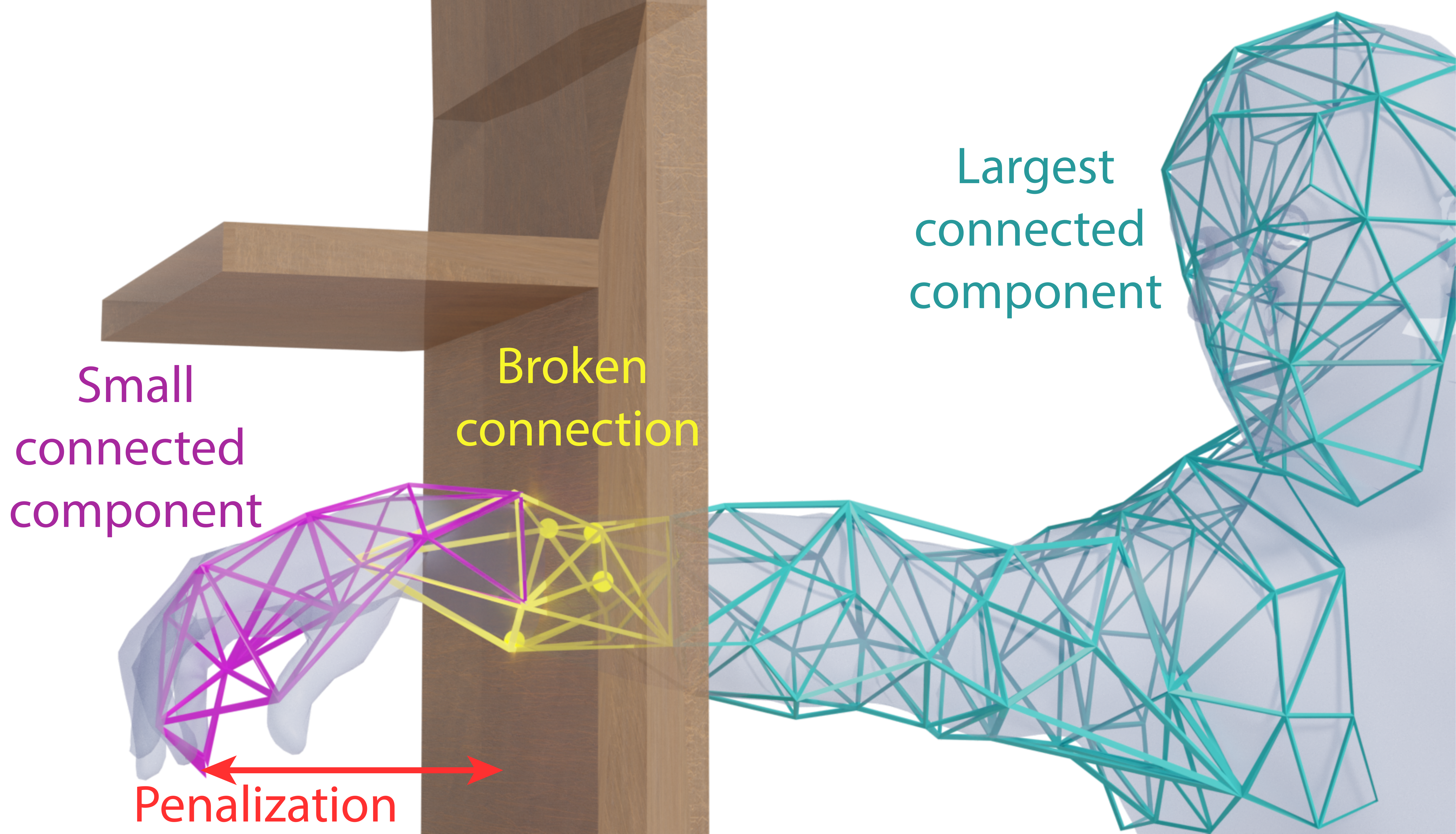}
    \caption{\textbf{Body connectivity}. When a body crosses an obstacle, the obstacle divides the body in two or more components, the break happening at the vertices that lie inside of the obstacle (in yellow). We penalize the vertices in all the connected components of the resulting body graph other than the largest one.}
    \label{fig:connected}
\end{figure}

\paragraph{Obstacle loss} To succeed at grasping objects in a scene, the humans that are generated need to avoid all the obstacles. We introduce a novel obstacle-avoidance loss that penalizes body-obstacle penetration, and consists of two parts: body-obstacle intersection and body mesh connectivity.

\textit{1. Body-obstacle intersection loss}. We represent obstacles as watertight meshes, and compute the signed distance from every vertex in $\mathcal{V}_{\text{b}}$ to the obstacle mesh: negative distances represent points \textit{inside} the obstacle. This loss sums the absolute values of the vertex-obstacle distances for all the vertices that lie inside the obstacle:
\begin{equation}
    \vspace{-0.1cm}
    \mathcal{L}^{\text{inter}}_{\text{o}} = \frac{1}{|\mathcal{V}_{\text{b}}|} \sum_{i=1}^{|\mathcal{V}_{\text{b}}|} \big|\min \left(0,d_{\text{vm}}(\mathcal{V}_{\text{b}_i}, \mathcal{M}_{\text{obstacle}}) \right)\big|,
    \label{eq:obst_loss}
\end{equation}
where $d_{\text{vm}}(.,.)$ is the signed distance function between a vertex (3D point) and a mesh.

\textit{2. Body mesh connectivity loss}. The previous loss alone does not penalize the parts of the body that penetrate the obstacle and resurface at the other side of the obstacle since these vertices are considered as ``outside'' vertices. To penalize them, we treat the human mesh as a graph. 
When part of the body penetrates an obstacle, two (or more) components of this graph get separated by the obstacle, resulting in a disconnected graph, composed of multiple connected components. The graph breaks precisely at the vertices that are inside of the obstacle (see~\cref{fig:connected}). Therefore, we penalize all the vertices that are not part of the largest connected components of the resulting graph, and the penalization is the distance from those vertices to the obstacle. Mathematically, the body mesh connectivity loss is:
\begin{equation}
    \mathcal{L}^{\text{con}}_{\text{o}} = \frac{1}{|\mathcal{V}_{\text{b}}|} \sum_{i=1}^{|\mathcal{V}^{\text{discon}}_{\text{b}}|} d_{\text{vm}}(\mathcal{V}^{\text{discon}}_{\text{b}_i}, \mathcal{M}_{\text{obstacle}}),
        \label{eq:con_loss}
\end{equation}
where the sum is over the set of vertices $\mathcal{V}^{\text{discon}}_{\text{b}}\subset \mathcal{V}_{\text{b}}$ that are outside of the obstacle mesh and belong to connected components of the resulting graph that are not the largest connected component, such as the pink one in~\cref{fig:connected}. For efficiency, this loss uses a subsampled version of $\mathcal{M}_{\text{b}}$. The final obstacle loss is the sum of the two: $\mathcal{L}_{\text{o}} = \mathcal{L}^{\text{inter}}_{\text{o}} + \mathcal{L}^{\text{con}}_{\text{o}}$.

\paragraph{Gaze loss} In addition to the main losses described above, we also incorporate an explicit gaze loss to encourage the human to look at the target object. We use the head direction vector from Taheri \textit{et al.}~\cite{taheri2021goal} which goes from the back to the front of the head $V_{b\rightarrow~f}$. Then we compute the vector from the back of the head to the target object $V_{b\rightarrow~o}$. The gaze loss minimizes the angle between these two vectors:
\begin{align*}
    \mathcal{L}_{\text{g}} &= \cos^{-1} \frac{V_{b\rightarrow~f} \cdot V_{b\rightarrow~o}}{|V_{b\rightarrow~f}||V_{b\rightarrow~o}|}
    \numberthis \label{eqn_rhmatch}
\end{align*}

During inference, we minimize a combination of the three losses:
    $\mathcal{L} = \lambda_{\text{hm}}\mathcal{L}_{\text{hm}} + \lambda_{\text{o}}\mathcal{L}_{\text{o}} + \lambda_{\text{g}}\mathcal{L}_{\text{g}}$.

\begin{figure}[t!]
    \centering
    \includegraphics[width=1.\linewidth]{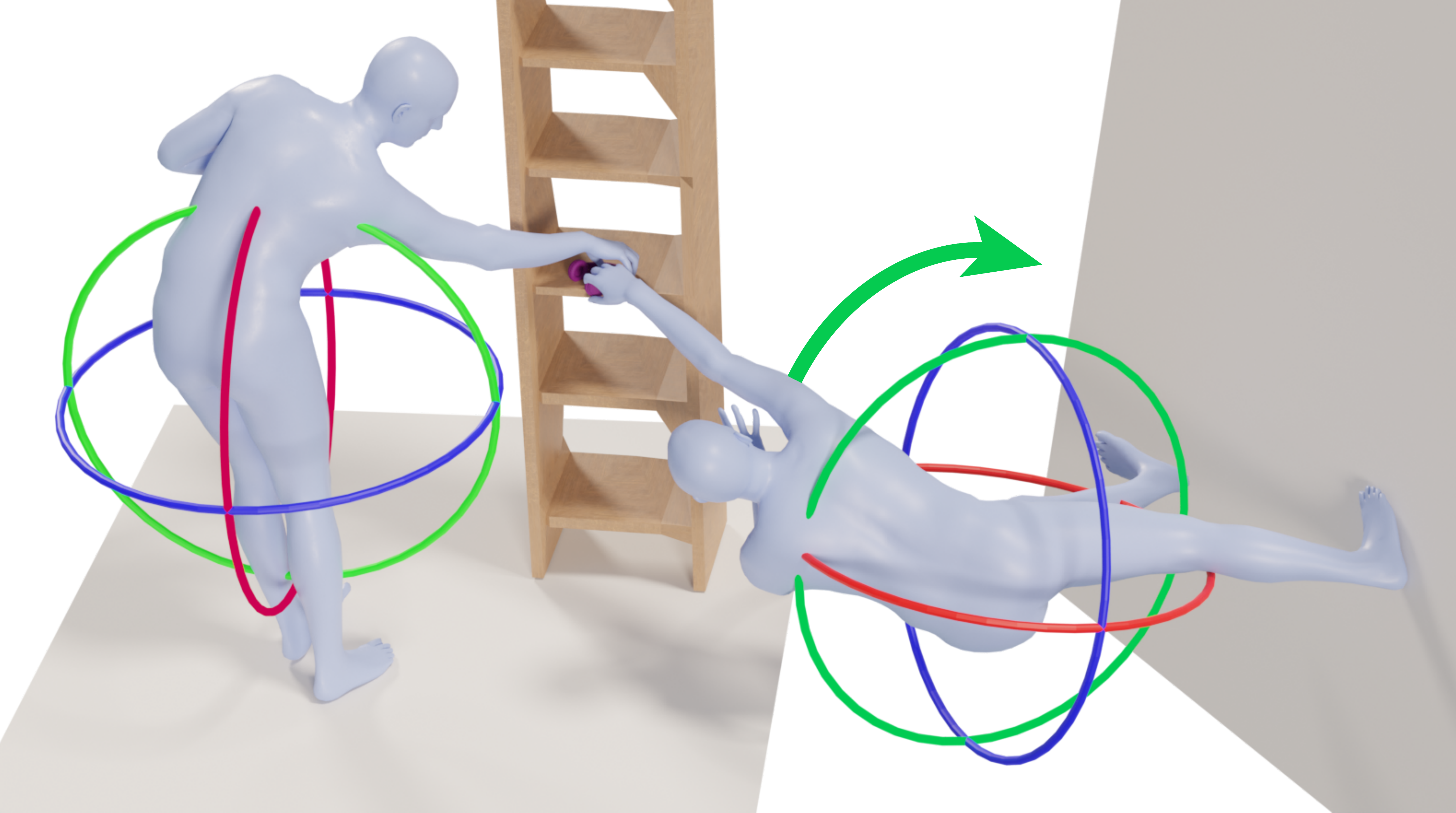}
    \caption{\textbf{Pose-ground prior}. Given a body pose, the position of the ground can be predicted. This determines the \dullgreen{pitch} and the \textcolor{red}{roll} of the body orientation, removing two degrees of freedom from our optimization. In this example, the human on the right should rotate to their right (pitch) and slightly forwards (roll) to get to a correct orientation with respect to the ground.}
    \label{fig:pgp}
\end{figure}

\begin{figure*}[t!]
    \centering
    \vspace{-0.5cm}
    \includegraphics[width=1.\linewidth]{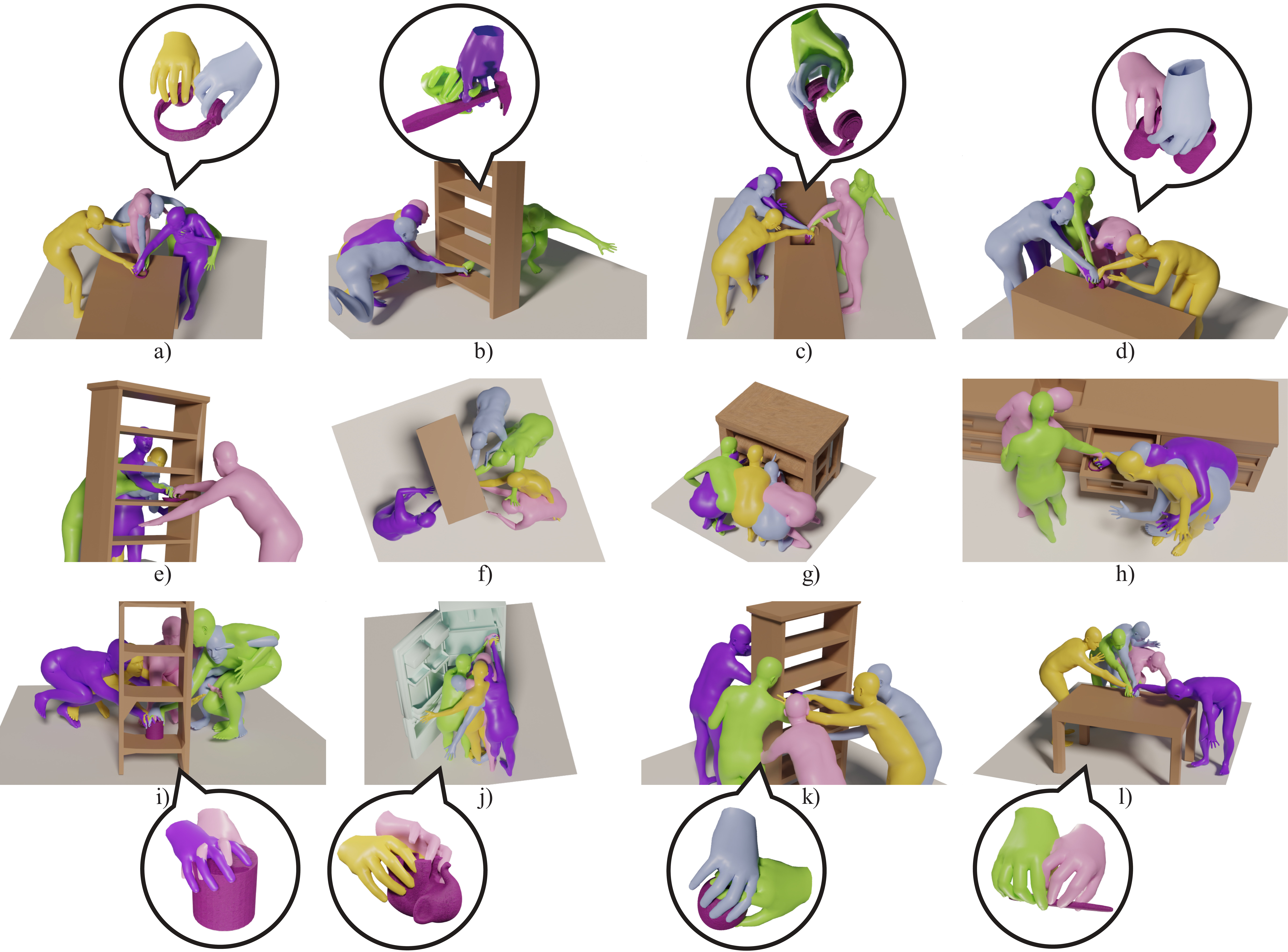}
    \caption{\textbf{\flex ibility}. We showcase \flex's ability to generate a variety of diverse, feasible full-body grasps in a number of challenging scenarios. Different colors indicate the top-5 generations. Bubbles zoom in on diverse hand grasps.}
    \label{fig:diversity}
\end{figure*}

\subsection{Pose-Ground Prior} 
Enforcing the previous losses can lead to perfect grasps, but potentially ``flying'' humans. 
To ensure that the human pose is also reasonable with respect to the ground, we learn a prior over the relationship between human pose and ground location.
Specifically, we use the AMASS dataset~\cite{AMASS:2019,mojo_zhang2021we}, which uses the \emph{xy}-plane as the floor, to train a 2-layer MLP that predicts the roll and pitch components of the human pelvis orientation with respect to the floor, given a human pose $\theta_{\text{b}}$. This MLP is trained using MSE regression, and it is kept frozen during inference. As exemplified in~\cref{fig:pgp}, given a specific human pose, the only actual degree of freedom for the human to have a natural orientation with respect to the ground is the yaw (shown in blue in the figure). The pitch and roll are constrained by the position of the ground.

Additionally, we fix the vertical component of the human body translation $t_{\text{b}}^z$ such that the lowest vertex of the predicted body mesh touches the ground.\\

\input{comparison}

\subsection{One Latent to Rule Them All}
\vspace{-.1cm}
Our framework requires optimizing the values of several interdependent parameters---for example, the angle of the hand grasp $\alpha$ constrains the human body orientation $R_{\text{b}}$ and pose $v$. Making sense of these relationships is required in order to smoothly search over the different parameters. Therefore, we delegate the burden of controlling the multiple parameters to a single controllable latent vector $z$ that can abstract away the low-level dependencies of the individual latent variables.
Specifically, following Liu \textit{et al.}~\cite{liu2022landscape}, we learn a mapping network (MLP in~\cref{fig:method}) from the latent vector $z$ to the different parameters and latents we defined above. At inference time, we optimize the values of $z$ and the weights of the MLP. The rest of the parameters are given as activations (outputs) of the MLP.

For every example (scene and object), we optimize $N$ latent vectors ($z$), with different initializations, and at the end of the optimization process, we select the ones that result in the smallest loss. The parameters of the mapping network (a 2-layer MLP) are shared across the $N$ latent vectors.
See \Cref{supp_training_details} for more implementation details.

%% file: comparison.tex
\newcommand{\tablestyle}[2]{\setlength{\tabcolsep}{#1}\renewcommand{\arraystretch}{#2}\centering\footnotesize}
\newcolumntype{Y}{>{\centering\arraybackslash}X}

\begin{table*}[t!]
\centering
\footnotesize
\tablestyle{2.5pt}{1.05}
\begin{tabularx}{1.0\linewidth}{lYYYYYY@{\hspace{2em}}YYYYY}
    \toprule
    & \multicolumn{6}{c}{\textbf{ReplicaGrasp}} & \multicolumn{5}{c}{\textbf{GRAB}} \\
    \cmidrule(lr{2.8em}){2-7} \cmidrule(lr){8-12}
     \multicolumn{1}{c}{\textbf{Method}} & \textbf{Obj Cont (\%) $\uparrow$} & \textbf{Obj Penet (\%) $\downarrow$} & \textbf{Obst Penet (\%) $\downarrow$} & \textbf{$\text{Div}_{\text{samp}}$ (cm) $\uparrow$} & \textbf{$\text{Div}_{\text{all}}$ (cm) $\uparrow$} & \textbf{Ground (cm) $\downarrow$} & \textbf{Obj Cont (\%) $\uparrow$} & \textbf{Obj Penet (\%) $\downarrow$} & \textbf{$\text{Div}_{\text{samp}}$ (cm) $\uparrow$} & \textbf{$\text{Div}_{\text{all}}$ (cm) $\uparrow$} & \textbf{Ground (cm) $\downarrow$} \\
    \midrule
    GOAL~\cite{taheri2021goal} & 1.14 & 2.14 & 6.87 & 0.20 & 37.89 & 5.31 & 1.62 & 3.50 & 0.11 & 7.87 & 3.56 \\
    SAGA~\cite{wu2022saga} & 1.21 & \textbf{0.29} & 7.27 & 1.10 & 43.84 & 9.40 & \textbf{2.19} & 3.43 & 1.06 & 17.35 & 2.39 \\
    \flex~(ours)& \textbf{2.20} & 2.49 & \textbf{0.53} & \textbf{10.37} & \textbf{69.78} & \textbf{0.00} & 1.63 & \textbf{2.73} & \textbf{24.94} & \textbf{46.46} & \textbf{0.00} \\
    \bottomrule
\end{tabularx}
\caption{\textbf{Results for ReplicaGrasp and GRAB}. We report contact and penetration with the object, obstacle penetration, diversity and distance from ground. \flex~almost always outperforms baselines, even though it was not trained on fully-body grasp data (GRAB).}
\label{tab:quant}
\end{table*}

%% file: 6_experiments.tex
\section{Experiments}
\label{sec:experiments}

We conduct experiments on two datasets: our challenging 1) ReplicaGrasp and 2) GRAB~\cite{GRAB:2020}. GRAB is a MoCap dataset of humans interacting with everyday objects without any other obstacles. Since baselines are trained and evaluated on GRAB, we evaluate if \flex~can perform comparably despite it not using GRAB's full-body grasping data.

\subsection{Baselines}
We compare with the only two existing works that perform full-body human grasping: GOAL~\cite{taheri2021goal} and SAGA~\cite{wu2022saga}.
Both methods use a conditional variational auto-encoder (cVAE)~\cite{kingma2013auto} to reconstruct 3D humans conditioned on the object's position and orientation, and have been trained on the full-body grasps of GRAB.
GOAL reconstructs a sub-sampled set of body vertices, while SAGA reconstructs surface body markers.
Both use test-time optimization for refining the full-body grasp by leveraging fine-grained human–object contact information.
Since these methods do not work when the objects are out of distribution in the horizontal \textit{xy}-plane, we evaluate them for objects placed at $(x,y)=(0,0)$. We then translate the resulting human-object pair to the correct $(x,y)$ for visualization.

\subsection{Ablations}
To understand the impact of the key elements of our approach, we perform ablation studies on a validation set of ReplicaGrasp containing 500 random object configurations.
We freeze the main optimization parameters independently: 1) hand-grasp latent $\theta_{\text{h}}$, and 2) human pose latent $\theta_{\text{b}}$. 
Additionally, we perform ablations to study the effect of: 3) removing the obstacle loss, and 4) not enforcing a pose-ground prior.

\subsection{Metrics}
\label{metrics}
Similar to prior work~\cite{taheri2021goal} we report object-grasping metrics, and additionally report obstacle and ground metrics:
\begin{itemize}[nosep,leftmargin=*]
\item \textbf{Object contact percentage} - percentage of object vertices in contact with the human mesh, 
\item \textbf{Object penetration percentage} - percentage of object vertices penetrated by the human mesh.
\item \textbf{Obstacle penetration percentage} - percentage of human vertices penetrating the obstacle mesh. 
\item \textbf{Ground distance} - absolute vertical distance from the lowermost vertex of the human mesh to the ground plane.
\end{itemize}

Higher object contact implies a more stable grasp, while lower obstacle and object penetration is naturally more preferred. We only compute object penetration for instances with non-zero contact.  Note that these metrics are not perfect -- e.g., lower contact could very well lead to a stable grasp. However, these are reasonably accepted proxies for automatically evaluating grasps \cite{GRAB:2020,taheri2021goal,wu2022saga,jiang2021graspTTA}. 

To evaluate the ability of different methods to generate diverse full-body grasps, following~\cite{wu2022saga} we also report:
\begin{itemize}[nosep,leftmargin=*]
\item \textbf{Sample diversity} - average vertex-to-vertex L2 distance for each sampled pair of human meshes for a single instance, averaged across instances. 
\item \textbf{Overall diversity} - average pairwise diversity across all pairs of generated samples for all instances of the dataset, which quantifies the method's ability to generate a range of complex human poses. 
\end{itemize}

%% file: 7_results.tex
\vspace{-0.1cm}
\section{Results}
\label{sec:results}

\subsection{Comparison to Baselines} 
\noindent~The main results comparing our method ~\flex~to the two baselines (GOAL~\cite{taheri2021goal} and SAGA~\cite{wu2022saga}) are shown in~\cref{tab:quant}. 

\begin{figure*}[t!]
    \centering
    \includegraphics[width=1.\linewidth]{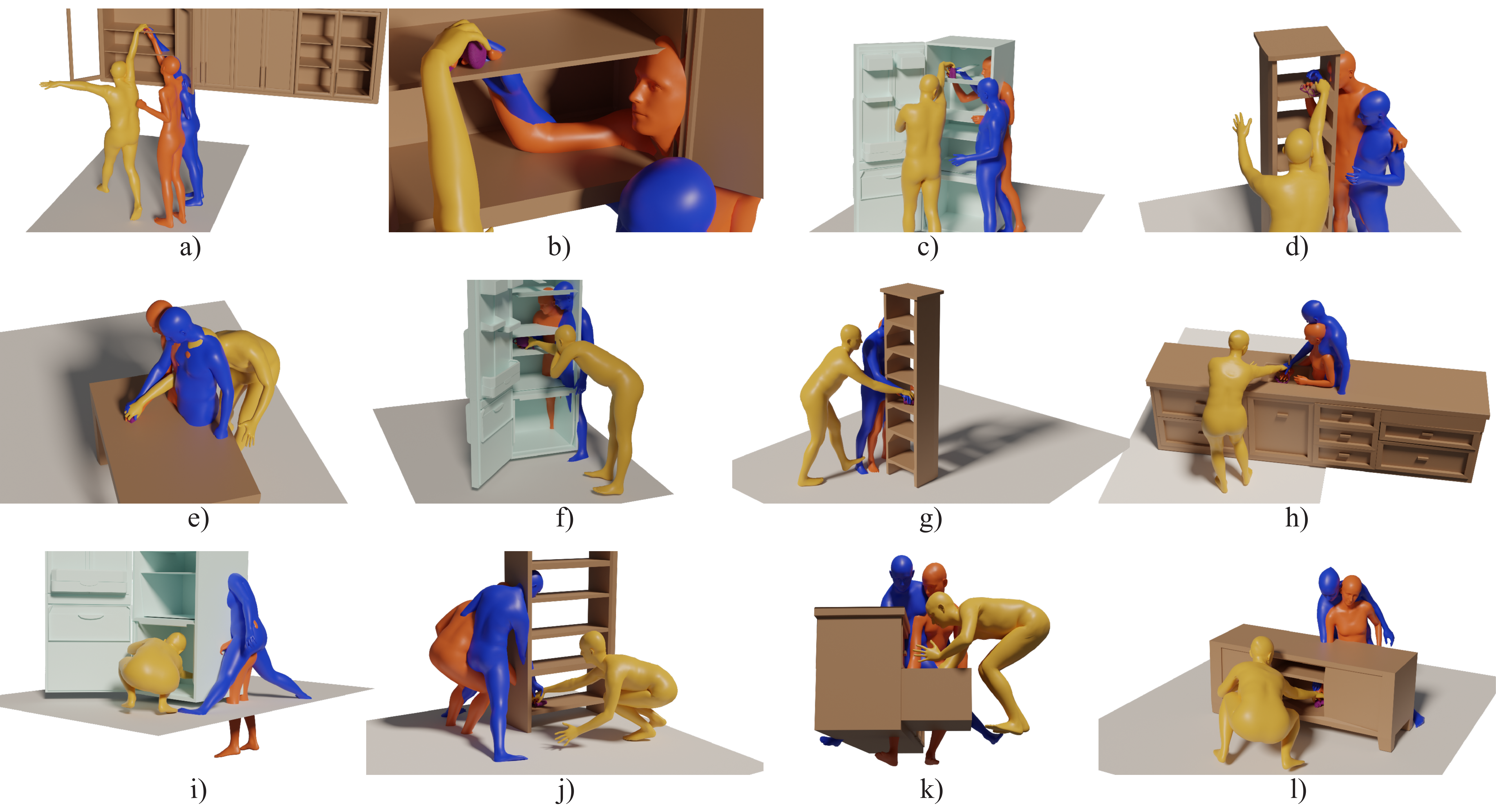}
    \caption{\textbf{Qualitative Comparisons}. We show 3D human avatars generated by \flexyellow{\flex}, \sagaorange{SAGA} and \goalblue{GOAL} when objects in the ReplicaGrasp dataset are placed at high \textbf{(top)}, medium \textbf{(middle)} and low \textbf{(bottom)} heights. \flexyellow{\flex} generates a variety of reasonable poses for grasping the target objects in a number of scenarios, while \sagaorange{SAGA} and \goalblue{GOAL} fail to generalize especially at extreme heights.}
    \label{fig:comparison}
\end{figure*}

\paragraph{\flex~generates diverse full-body grasps, even with obstacles}
On ReplicaGrasp, which tests grasping objects in more realistic scenarios, ~\flex~achieves the best object contact score. This is because in many cases where the object is either too high or too low, the baselines often fail to even touch the object.
We observed that SAGA performs best on object penetration. This is explained by an explicit optimization to minimize inference-time collision loss for \emph{1500 iterations}. However, this procedure may lead to unnatural humans -- for example, humans that penetrate the ground to grasp lower objects (see~\cref{fig:comparison} i,k) or elongated humans for grasping higher objects (see~\cref{fig:comparison} a). This phenomenon is reflected in the ground distance metric, where SAGA generates humans on average 9 cm away from the ground. 
Instead,~\flex, which combines the full-body, hand-grasping, and pose-ground priors, all of which are data-driven, will be in-distribution with the data by construction.
Regarding diversity,~\flex~outperforms baselines by a significant margin, while also avoiding obstacles. 

\paragraph{GRAB} On the GRAB dataset,~\flex~performs comparably to both baselines on the object contact and penetration metrics. Note that~\flex~achieves similar performance (surpassing GRAB on both metrics, and SAGA on one) \emph{without using any full-body grasping data}. Additionally,~\flex~surpasses both baselines by large margins on diversity metrics.

\input{ablations}

\subsection{Ablations}
We show the importance of different components of \flex~by demonstrating how performance is negatively impacted when each is removed in \cref{tab:ablations}.

\begin{figure}[t!]
    \centering
    \includegraphics[width=1.\linewidth]{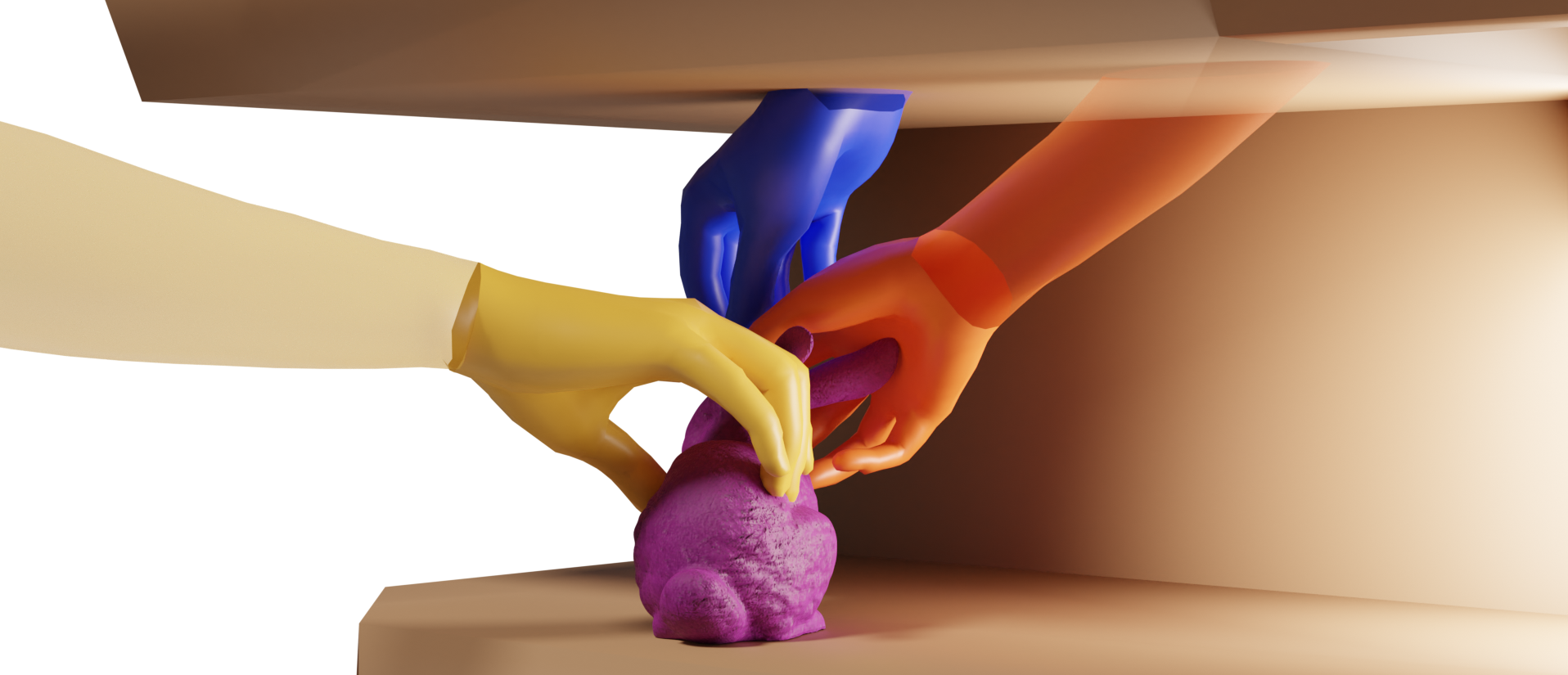}
    \caption{\textbf{Hand grasp search.} We illustrate the benefit of searching in the latent space of the hand grasping model. Hallucinating the full body associated with each grasp makes the choice easier.}
    \label{fig:hand_search}
\end{figure}

\noindent \textbf{-- No hand grasp optimization leads to the worst hand grasp with minimal contact.} When we do not allow search in the latent space $w$ of the hand-grasping model (Row 2), we get a poor grasp. Freezing the hand grasp severely restricts the space of feasible poses, leading to a deadlock between the hand-matching and obstacle losses. Hence, avoiding obstacles leads to the object being touched minimally. We further demonstrate the importance of searching for hand-grasps in \cref{fig:hand_search}. While all the hand-grasps are plausible, upon hallucinating the full-body human accompanying the grasp, it is evident that the human will penetrate the scene for the blue and orange grasps. Thus, searching over hand grasps will allow the model to pick grasps that lead to minimal object penetration.

\noindent \textbf{-- No body pose optimization leads to high obstacle penetration and lower diversity.}
By not searching over the full body latent space $v$, the model is incapable of \textit{\flex ibly} adjusting the human pose to avoid obstacles and thus generates humans that intersect with them. For the same reason, there is little variation in overall full-body grasp predictions as evident from the low diversity numbers.

\noindent \textbf{-- No obstacle loss leads to high obstacle penetration.}
Comparing Row 4 with Row 1, we see that removing the obstacle loss (\cref{eq:obst_loss,eq:con_loss}) leads to very high obstacle penetration. Because humans are not incentivized to avoid obstacles, there are more feasible body poses, which explains the significant increase in diversity.

\noindent \textbf{-- Removing pose-ground prior results in flying humans.}
Removing the pose-ground prior (Row 5) no longer forces humans to touch the ground, resulting in a very large distances ($71.7$ cm) from the ground.

\subsection{Qualitative Results}
\vspace{-0.3cm}
\looseness=-1
\paragraph{Diversity} \cref{fig:diversity} showcases \flex's ability to discover a variety of reasonable grasps while simultaneously satisfying obstacle constraints. For example, in~\cref{fig:diversity}~d,l),~\flex~ generates both crouched and standing humans in different positions and orientations without penetrating the obstacle or compromising on the quality of the right hand grasp. When the object is very low, as in~\cref{fig:diversity}~b),~\flex~generates squatting as well as kneeling positions. 

\paragraph{Comparison to baselines} Qualitative results for~\flex, SAGA and GOAL are shown in~\cref{fig:comparison} in \flexyellow{yellow}, \sagaorange{orange} and \goalblue{blue} colored humans, respectively. GOAL's grasps lack diversity -- the right hand always grasps the object from the top with very little variation in the distance to the object as well as the overall pose. When the object is too low, it starts to generate unnatural legs (see~\cref{fig:comparison} i,j,k).

SAGA generates more diverse hand grasps (side and bottom grasps in~\cref{fig:comparison} c,d,h) and full-body poses (slightly bent legs in~\cref{fig:comparison} j). However, it often fails when objects are too low or too high (see~\cref{fig:comparison} a,i,k).

\flex~is able to generate suitable poses based on the object placement with respect to the scene. For example, when the object is deep inside the refrigerator (see~\cref{fig:comparison}~f,i), \flex~can generate humans with a carefully outstretched hand such that it doesn't collide with any obstacles. In~\cref{fig:comparison}~a), where the object is very high on the shelf, \flex~generates a human on their toes while in~\cref{fig:comparison}~c), \flex~generates a good right hand grasp that carefully avoids the protruding component in the top compartment of the fridge. \cref{fig:comparison} e) demonstrates how~\flex~can match the hand grasp of the baselines, and additionally jut out the rest of the body to avoid the obstacle when necessary.

\begin{figure}[t!]
    \centering
    \includegraphics[width=1.\linewidth]{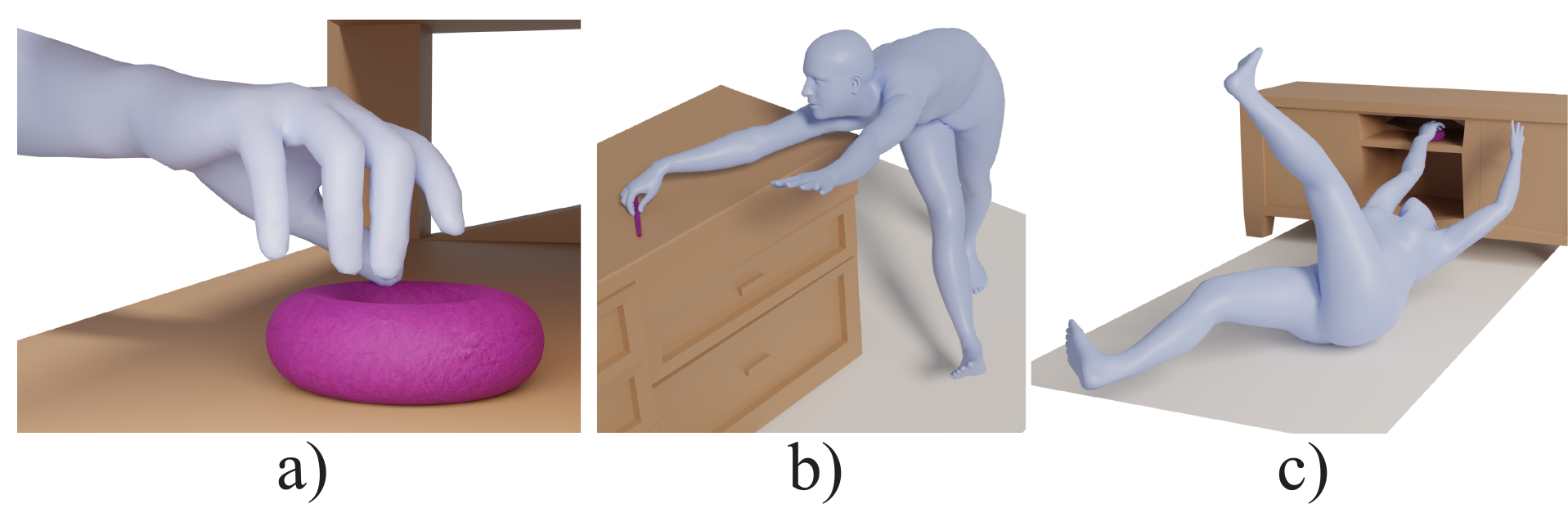}
    \caption{\textbf{Failure cases}. \textbf{a)} Our selected grasping model $\mathcal{G}$ fails. \textbf{b)} All our constraints are satisfied, but the human pose is too stretched / unnatural. \textbf{c)} The pose-ground prior may be imperfect or result in rare poses while satisfying other constraints.}
    \label{fig:failure}
\end{figure}

\paragraph{Failures} \cref{fig:failure} shows some failures. Most of them are caused by the limitations of the pre-trained generative models $\mathcal{G}$ and $\mathcal{P}$. Fortunately, \flex~is model agnostic, so it will only improve as better generative models are developed. \flex~only requires them to be differentiable.

\begin{SCfigure}
    \centering
    \includegraphics[width=0.6\linewidth]{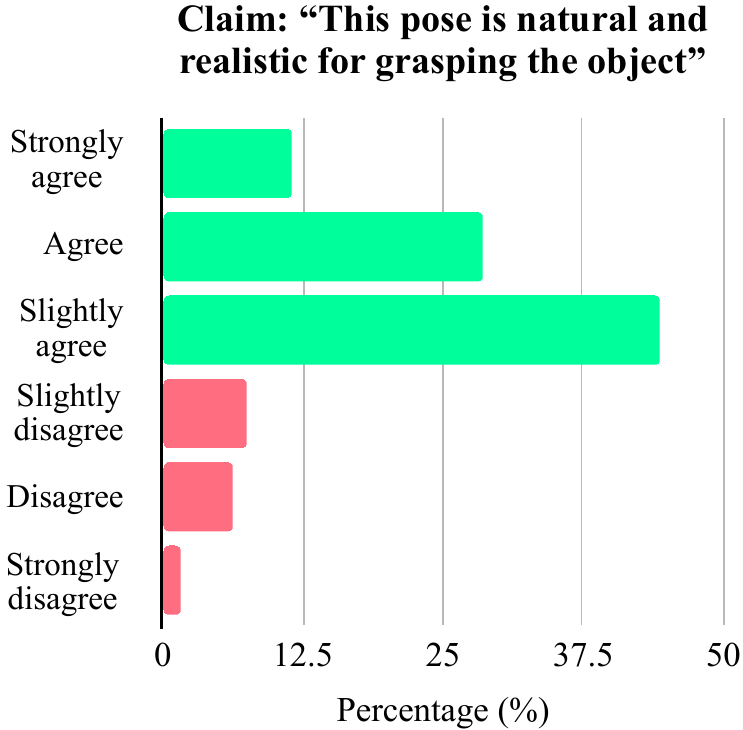}
    \caption{\textbf{Human Studies} We conduct human studies on Amazon Mechanical Turk and ask subjects to rate 3D human grasps generated with \flex~on a Likert scale of 1-6 for realism. Subjects agree 85\% of the time that our generated humans grasp objects realistically.}
    \label{fig:amt}
\end{SCfigure}

\subsection{Human Evaluation}
To evaluate the perceptual quality of our generated full-body grasps, we conducted human studies on Amazon Mechanical Turk (AMT). We chose a subset of \flex~results covering all 48 receptacles twice for objects in both fallen and upright orientations. We showed each result to five different subjects on AMT in an interactive 3D interface and asked them to rate the full-body grasps on a Likert scale of 1 (strongly disagree) to 6 (strongly agree). We filtered out responses with standard deviation $>1$.
Results are shown in Fig.~\ref{fig:amt}. 85\% of the time participants agree that our generations are natural and realistic.

A subject who gave a rating of 2 wrote: \textit{``[...] arm that is grabbing is reaching way too hard to reach object, but the grasp is natural''}, while a subject who gave a rating of 6 wrote: \textit{``The bend on the spine and the legs being approximate shoulder width apart is a natural body movement''}.

%% file: ablations.tex
\begin{table}[t!]
\centering
\footnotesize
\tablestyle{2.5pt}{1.05}
\begin{tabularx}{1.0\linewidth}{clYYYYYY}
    \toprule
    \# & \multicolumn{1}{l}{\textbf{Method}} & \textbf{Obj Cont (\%) $\uparrow$} & \textbf{Obst Penet (\%) $\downarrow$} & \textbf{Ground (cm) $\downarrow$} & \textbf{$\text{Div}_{\text{samp}}$ (cm) $\uparrow$} & \textbf{$\text{Div}_{\text{all}}$ (cm) $\uparrow$} \\
    \midrule
    1 & Ours & 2.19 & 0.97 & 0.00 & 10.15 & 70.44 \\
    \midrule
    2 & No grasp optimization & \red{0.88} & 0.85 & 0.00 & 13.19 & 71.07 \\
    3 & No pose optimization & 2.03 & \red{1.87} & 0.00 & \red{0.27} & \red{41.55}\\
    4 & No obstacle loss & 2.38 & \red{10.57} & 0.00 & {21.60} & 71.86 \\
    5 & No pose-ground prior & 2.13 & 0.61 & \red{71.72} & 16.88 & 96.27  \\
    \bottomrule
\end{tabularx}
\caption{\textbf{Ablation Studies of different components~\flex}. For each metric, \red{red} represents a significant drop.}
\label{tab:ablations}
\end{table}

%% file: 8_conclusion.tex
\section{Conclusion}
\label{sec:conclusion}
In this work, we address the task of generating full-body humans grasping 3D objects in the presence of obstacles and introduce a new dataset, ReplicaGrasp, to evaluate the realism of the generations.
We describe an optimization-based approach that leverages existing hand-grasping models and human pose priors to solve this task, without using any 3D full-body grasping data.
Experiments show that we are able to generate realistic avatars that surpass existing methods, both quantitatively and qualitatively.

\textbf{Acknowledgements:} This research is based on work partially supported by NSF NRI Award \#2132519, and the DARPA MCS program under Federal Agreement No. N660011924032. D.S.\ is supported by the Microsoft PhD fellowship. The views and conclusions contained herein are those of the authors and should not be interpreted as necessarily representing the official policies, either expressed or implied, of the sponsors.

%% file: X_supplementary.tex
\appendix

\newpage

\twocolumn[
\centering
\Large
\textbf{Appendix} \\
\vspace{1.0em}
]
\appendix

\section{Implementation details}
\label{sec:supp_impl_details}

\subsection{Pre-trained Generative Models}

\paragraph{Full-body pose prior} For our full-body generative model $\mathcal{P}$, we use the \href{https://github.com/nghorbani/human_body_prior}{GitHub} implementation of VPoser \cite{SMPL-X:2019}.
VPoser is a variational autoencoder~\cite{kingma2013auto} trained on data obtained by applying MoSh~\cite{MoSh_lopermahmoodetal2014} on three publicly available human motion capture datasets: CMU~\cite{cmuWEB}, training set of Human3.6M~\cite{human36m}, and the PosePrior dataset~\cite{PosePrior_Akhter:CVPR:2015}.

\paragraph{Hand object grasping model} For the right-hand grasping model $\mathcal{G}$, we use the \href{https://github.com/otaheri/GrabNet}{GitHub} implementation of GrabNet \cite{GRAB:2020}. 
GrabNet consists of two networks: \textbf{1) CoarseNet} for coarse grasps and \textbf{2) RefineNet} for refining the coarse grasps. 
CoarseNet takes as input a latent vector $w$, and the object $\mathcal{O}$, represented by its BPS representation~\cite{bps_prokudin2019efficient}, and generates a hand pose, including its translation and rotation, to be used as input to MANO. 
RefineNet takes as input the CoarseNet grasp and the distances $D$ from the coarse MANO vertices to the object mesh. It then refines the grasp through 3 iterations as in~\cite{kanazawa2018end} to give the final grasp.
RefineNet has been trained by sampling CoarseNet grasps as ground truth and perturbing the hand pose parameters to simulate noisy input estimates. 

\paragraph{Object representation} 
The object to-be-grasped $\mathcal{O}$ is represented in GrabNet using the Basis Point Set (BPS)~\cite{bps_prokudin2019efficient} representation which is capable of encoding arbitrary 3D object shapes. 
Given any object vertices, an approaching angle $\alpha$ and $N_b$ fixed basis points, the BPS representation involves rotating the object by the angle $\alpha$, placing it in the center of the points and calculating the minimum distance from each point to the nearest surface of the object. The outputs of our model are rotated by the inverse of the rotation matrix given by $\alpha$.
We use the implementation from the bps\_torch library on \href{https://github.com/otaheri/bps_torch}{GitHub}.

\subsection{Training Details}
\label{supp_training_details}
\begin{itemize}[nosep,leftmargin=*]
\item For every example (scene and object), we optimize $N=500$ latent vectors $z$, with different initializations, and at the end of the optimization process we select the ones that result in the smallest loss. During training, we periodically discard the 50\% of the latent vectors that produce the largest losses. At the end of the optimization process, we end up with the best 16 samples out of the 500. The parameters of the mapping network (consisting of a 2-layer MLP) are shared across the $N$ latent vectors.

\item We constrain the value of $w$ by normalizing it such that its norm is always one, following the density of a high-dimensional Gaussian prior, thus making sure $w$ is within the distribution of the latent space of $\mathcal{G}$.

\item We train with Adam optimizer with a learning rate of $1e-3$ for $z$, and $1e-4$ for the mapping network. Additionally, we found that the translation parameters have a much stronger gradient than the rest, especially the latent $v$, so we divide the gradient that goes through $t_b^{xy}$ by 3, and multiply the gradient that flows through $v$ by 10. We train for 500 iterations.

\item Empirically, we found that the pre-trained GrabNet model was much more sensitive to approaching angle $\alpha$ then it was to the latent $w$, hence we set $w$ to the zero vector in our experiments.

\item For the hand matching loss, we weigh the vertices around the wrist more ($\times$3) than the rest, as the alignment around the wrist is less noisy than in the fingers.

\item The values for the loss weights $\lambda$ in the total loss are set to: $\lambda_{\text{hm}}=20$, $\lambda_{\text{o}}=1000$, $\lambda_{\text{g}}=0.01$. These do not necessarily reflect the importance given to each loss, as the loss values are in completely different scales.

\item We additionally found that scaling the output of the MLP differently for every parameter was helpful. Specifically, we scale $v$ by 5 (giving more flexibility to the human pose generator), the translation parameters by 10, and the angle and orientation by 20. 

\item We found that some obstacles have very thin walls, so we make the obstacle mesh $\mathcal{M}_{\text{obstacle}}$ thicker by 5mm, which allows us to model the intersections better.
\end{itemize}

\section{ReplicaGrasp Dataset}
\label{sec:supp_replicagrasp_dataset}
\paragraph{Receptacles} We use a total of 48 receptacles from the ReplicaCAD dataset~\cite{szot2021habitat}.
Some of the static rigid object receptacles include: apartment chair, sofa, table top, TV stand and wall cabinet.
The receptacles from articulated objects include: refrigerator top, middle and bottom; top, middle and bottom drawers of kitchen counter on both right and left sides, and kitchen sink; as well as top, middle and bottom compartments of both sides of the kitchen cupboard.
Many of the receptacles are visible in~\cref{fig:teaser}.

\paragraph{Objects} We obtain 50 everyday object meshes from the GRAB dataset and use the Habitat Simulator~\cite{habitat19iccv} to get the final locations of the objects on the receptacles. We use the \href{https://github.com/facebookresearch/habitat-sim}{GitHub v0.2.2 release}. The simulator runs dynamics for 5 seconds to check for stability of newly placed objects.

\begin{figure}[t!]
    \centering
    \includegraphics[width=1.\linewidth]{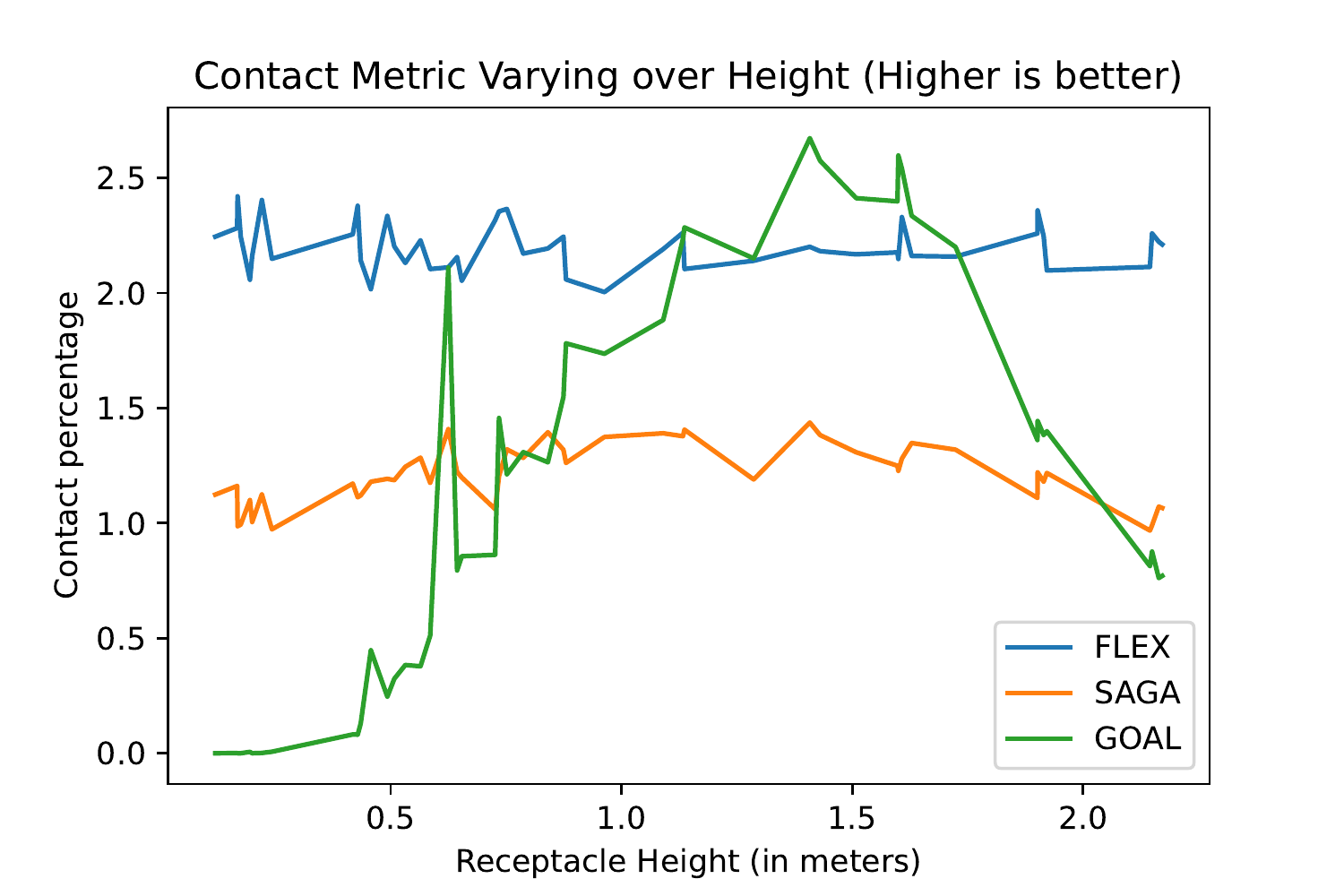}
    \caption{\textbf{Object contact percentage varying by height.}~\flex~performs more consistently than both baselines and is best on average.}
    \label{fig:contact}
\end{figure}

\begin{figure}[t!]
    \centering
    \includegraphics[width=1.\linewidth]{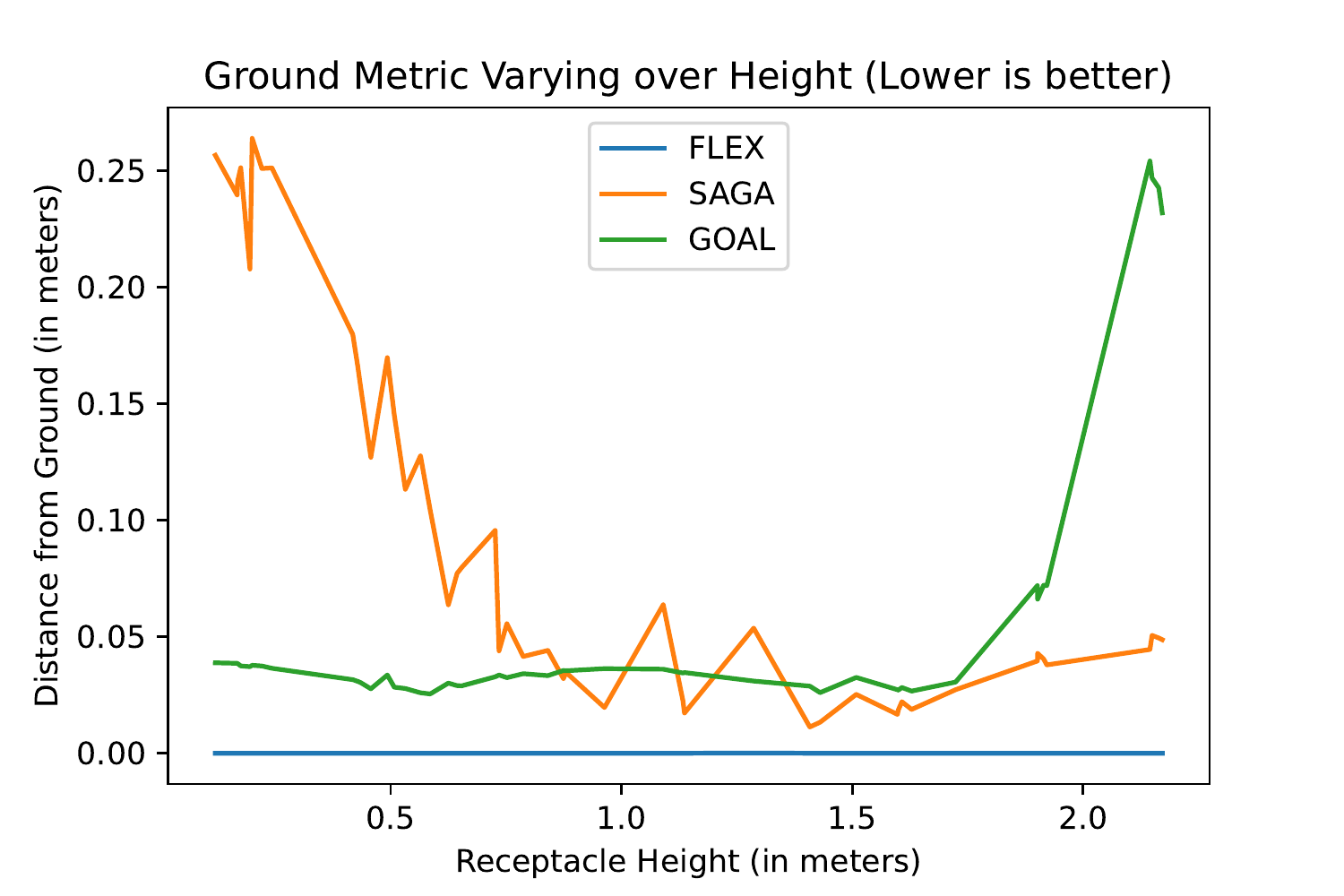}
    \caption{\textbf{Ground distance varying by height.} SAGA performs worst at lower heights while GOAL performs worst when the objects are high up.}
    \label{fig:ground}
\end{figure}

\begin{figure}[t!]
    \centering
    \includegraphics[width=1.\linewidth]{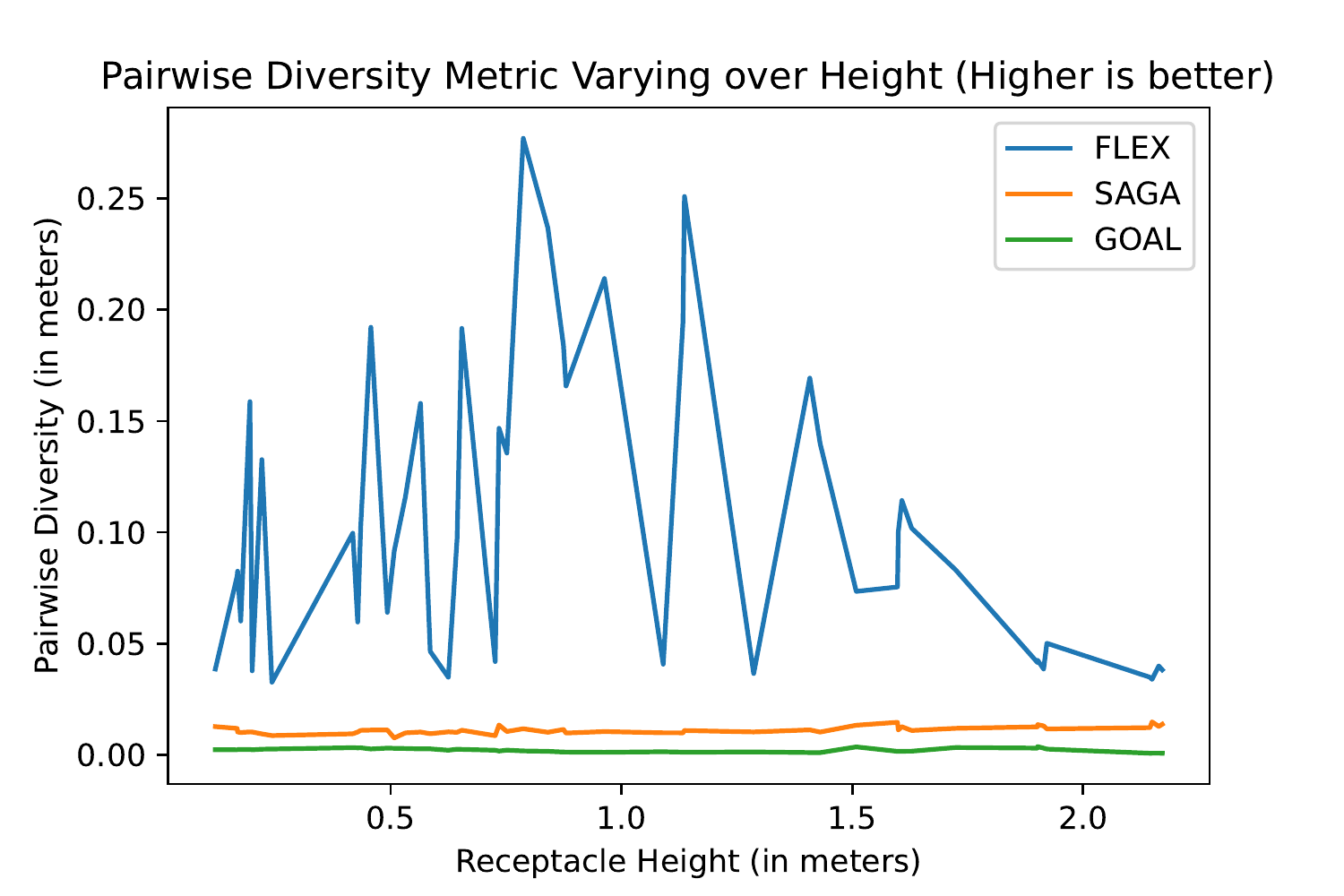}
    \caption{\textbf{Diversity varying by height.} GOAL and SAGA both consistently fail at generating diverse outputs at all heights. \flex~can generate most diverse grasps at medium heights.}
    \label{fig:div}
\end{figure}

\begin{figure}[t!]
    \centering
    \includegraphics[width=1.\linewidth]{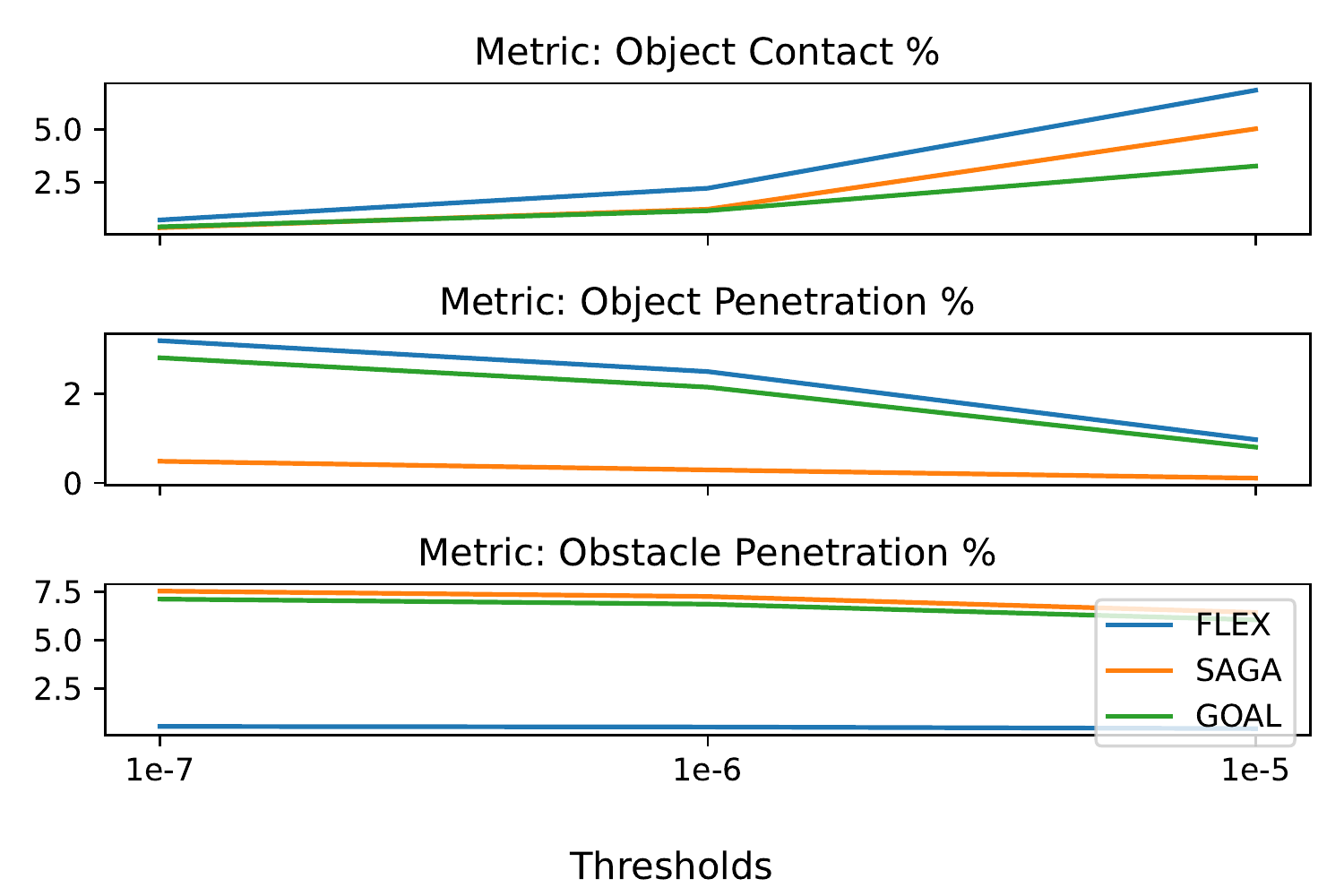}
    \caption{\textbf{Sensitivity to threshold $\sigma$.} Trends stay the same for all metrics across different thresholds.}
    \label{fig:sensitivity}
\end{figure}

\begin{figure*}[t!]
    \centering
    \includegraphics[width=1.\linewidth]{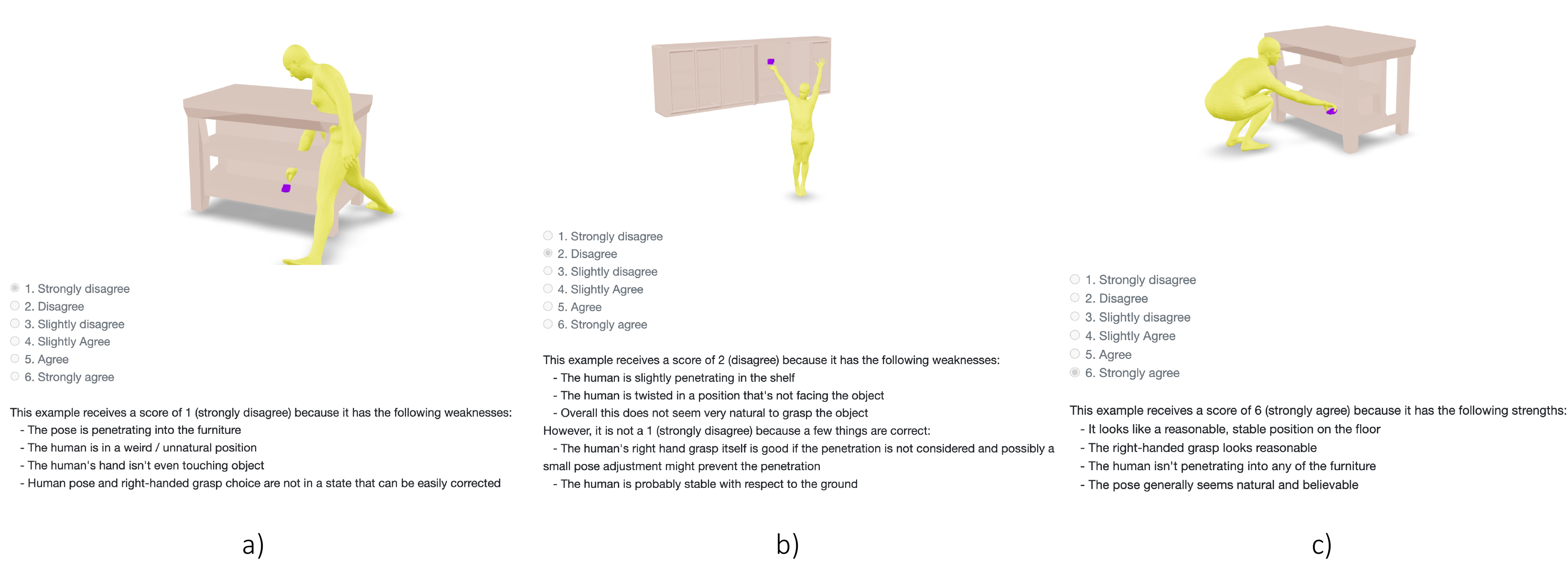}
    \caption{\textbf{Examples shown to Amazon Mechanical Turk subjects}. We provide these three examples to the subjects, which range from very bad (strongly disagree with the statement) to very good (strongly agree) results.}
    \label{fig:demo}
\end{figure*}

\begin{figure}[t!]
    \centering
    \includegraphics[width=1.\linewidth]{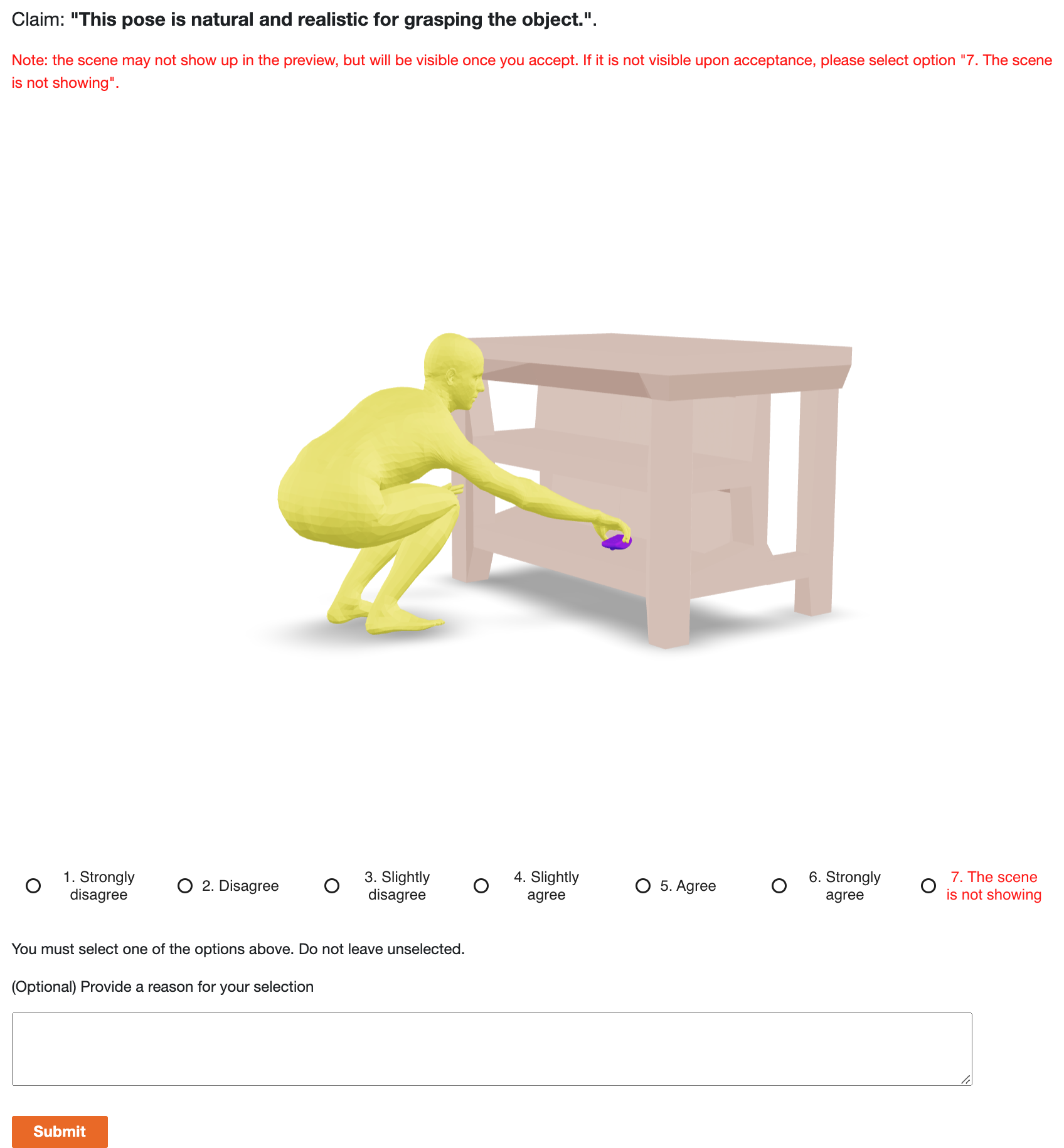}
    \caption{\textbf{Human studies.} Example of a HIT shown to subjects on Amazon Mechanical Turk.}
    \label{fig:mturk_prompt}
\end{figure}

\begin{figure*}[t!]
    \centering
    \includegraphics[width=1.\linewidth]{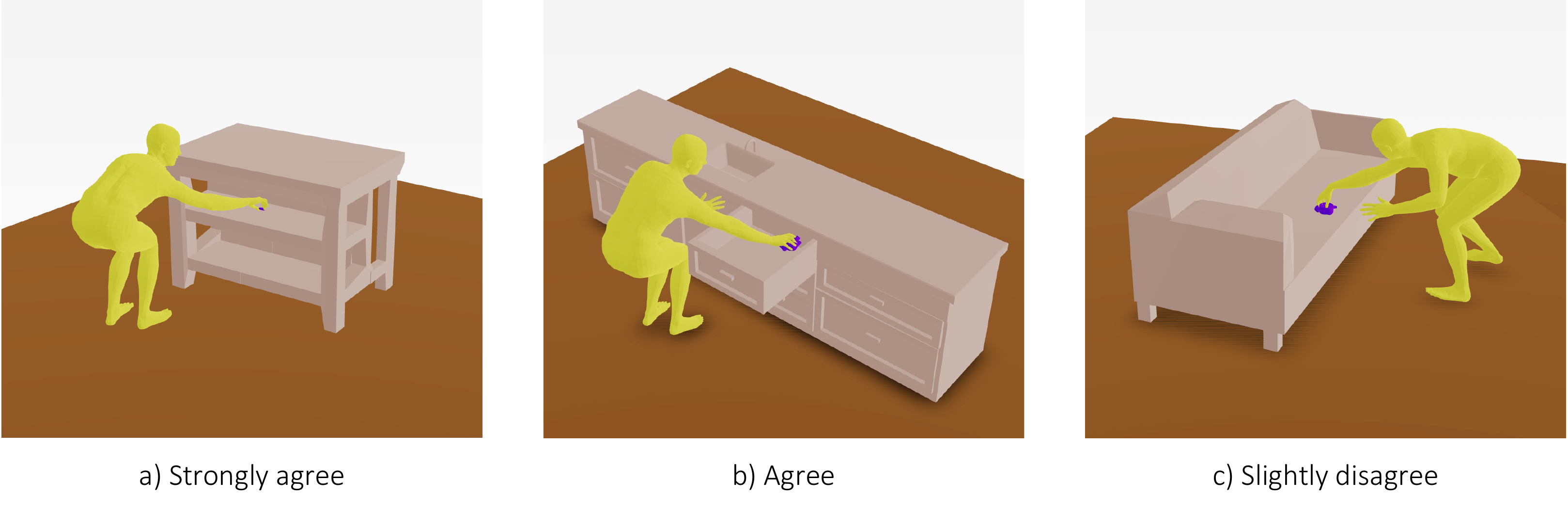}
    \caption{\textbf{Examples of human ratings.} The figure shows three samples and their corresponding human ratings. See \Cref{supp_human_studies} for comments from subjects.}
    \label{fig:mturk_viz}
\end{figure*}

\section{Quantitative Analysis}
\label{sec:supp_quant_analysis}

\noindent~We conduct a detailed analysis of the results in \Cref{tab:quant}.

\paragraph{Performance as a function of object height} We demonstrate the need of having a benchmark like ReplicaGrasp that allows evaluating grasps at different heights. 
In GRAB, the object heights varies from a minimum of 0.75 meters to a maximum of 1.38 meters, with a mean of 1 meter. In ReplicaGrasp, our object heights have a much larger range from a minimum of 0.12 meters to a maximum of 2.2 meters, with a mean of 1 meter.
We show how performance along different metrics changes by varying the heights of the objects.
\begin{itemize}[nosep,leftmargin=*]
\item \textbf{Object contact percentage} - \cref{fig:contact} shows that GOAL performs well when objects are at heights that have been seen during training, but sees drops in performance at other heights. SAGA is more consistent than GOAL even at varying heights. \flex~outperforms both baselines on average across all heights without showing much variation across height changes.
\item \textbf{Ground distance} - \cref{fig:ground} shows that when the object is at a low height, SAGA fails by generating humans with legs buried below the ground. SAGA is better at higher heights, although qualitatively the humans appear elongated.
GOAL generates humans that try to fly up to grasp objects at larger heights.
GOAL performs better at lower heights, although qualitatively the humans look unnatural with awkwardly bent legs.
\item \textbf{Sample diversity} - In \cref{fig:div}, we show the average pairwise diversity across pairs of samples generated for an instance, averaged across all instances of the dataset. This quantifies the method's ability to generate a range of complex human poses. 
\flex~outperforms both baselines by a large margin despite having additional constraints of avoiding obstacles.
\end{itemize}

\paragraph{Sensitivity of the metrics to the threshold} In order to compute the metrics, we set a threshold $\sigma$ that determines the boundary between contact and penetration with an object or an obstacle. In \cref{fig:sensitivity} we report results of our metrics for different values of this threshold, and show that the metrics are not sensitive to its value, and that the trends shown in the main paper (where we use $\sigma=1e-6$) hold.

\section{Evaluation Metrics}
\label{sec:supp_eval_metrics}
We provide equations for each of the metrics described in \Cref{metrics}.

\paragraph{Object contact percentage}
\begin{equation}
    M^{\text{contact}}_{\text{obj}} = \frac{100}{|\mathcal{V}_{\text{o}}|} \sum_{i=1}^{|\mathcal{V}_{\text{o}}|} \mathds{1}\big(
    |d_{\text{vm}}(\mathcal{V}_{\text{o}_i}, \mathcal{M}_{\text{human}})| \leq \sigma
    \big),
    \label{eq:obj_cont}
\end{equation}
where $\mathcal{V}_{\text{o}}$ are the vertices of the object, $\mathcal{M}_{\text{human}}$ is the human mesh, $d_{\text{vm}}$ is the signed vertex-to-mesh distance, $\sigma$ is a small threshold and $\mathds{1}$ is the indicator function.

\paragraph{Object penetration percentage}
\begin{equation}
    M^{\text{penet}}_{\text{obj}} = \frac{100}{|\mathcal{V}_{\text{o}}|} \sum_{i=1}^{|\mathcal{V}_{\text{o}}|} \mathds{1}\big(
    d_{\text{vm}}(\mathcal{V}_{\text{o}_i}, \mathcal{M}_{\text{human}}) < -\sigma
    \big)
    \label{eq:obj_penet}
\end{equation}

\paragraph{Obstacle penetration percentage}
\begin{equation}
    M^{\text{penet}}_{\text{obst}} = \frac{100}{|\mathcal{V}_{\text{b}}|} \sum_{i=1}^{|\mathcal{V}_{\text{b}}|} \mathds{1}\big(
    d_{\text{vm}}(\mathcal{V}_{\text{b}_i}, \mathcal{M}_{\text{obstacle}}) < -\sigma
    \big),
    \label{eq:obst_penet}
\end{equation}
where $\mathcal{V}_{\text{b}}$ are the vertices of the human body and $\mathcal{M}_{\text{obstacle}}$ is the obstacle mesh. In contrast to \cref{eq:obj_penet}, here we average over the human body (not the obstacle) vertices, because we care about how much of the human is penetrating an obstacle, not how much of the obstacle is being penetrated by the human.

\paragraph{Ground distance}
\begin{equation}
    M_{\text{ground}} =
    \big| \min (\mathcal{V}^z_{\text{b}}) \big|,
    \label{eq:ground}
\end{equation}
where $\mathcal{V}^z_{\text{b}}$ is the $z$ component of all vertices in $\mathcal{V}_{\text{b}}$.

\paragraph{Sample diversity}
\begin{equation}
    M_{\text{Div}_\text{samp}} =
    \frac{2}{\mathcal{N}_{\text{s}} \cdot (\mathcal{N}_{\text{s}}-1)} \sum_{\substack{i,j \in \mathcal{N}_{\text{s}} \\ i\neq~j}}{d_{\text{vv}}(\mathcal{V}^{(i)}_{b}, \mathcal{V}^{(j)}_{b})},
    \label{eq:samp_div}
\end{equation}
where $\mathcal{N}_{\text{s}}$ is the number of samples for a single example and $d_{\text{vv}}$ is the $L^2$ vertex-to-vertex distance in the 3D space. $\mathcal{V}^{(i)}_{b}$ represents the human body vertices corresponding to the $i$-th sample.

\paragraph{Overall diversity}
\begin{equation}
    M_{\text{Div}_\text{all}} =
    \frac{2}{\mathcal{N}_{\text{d}} \cdot (\mathcal{N}_{\text{d}}-1)} \sum_{\substack{i,j \in \mathcal{N}_{\text{d}} \\ i\neq~j}}{d_{\text{vv}}(\mathcal{V}^{(i)}_{b}, \mathcal{V}^{(j)}_{b})},
    \label{eq:overall_div}
\end{equation}
where $\mathcal{N}_{\text{d}}$ is the total number of instances in the dataset.

\section{Human Studies}
\label{supp_human_studies}
We conducted perceptual evaluation studies on Amazon Mechanical Turk (AMT) with a prompt as shown in \cref{fig:mturk_prompt}. The specific instructions were as follows:
\begin{lstlisting}[breakatwhitespace=true]
We want to evaluate the realism of the humans. Some questions to ask yourself while solving the task:
1) Would you expect to see a human like this in real-life?
2) Is the hand grasp going to result in a natural grasp?
3) Is the human stable on the ground?

The scene can be navigated by:
1) Clicking and dragging the mouse, to rotate the scene.
2) Zooming in and out with the scroll wheel.
3) Clicking at a point in the scene, to position that point in the center of the scene.

See examples for a better intuition.
\end{lstlisting}
Further, we showed subjects three examples of how to successfully perform the task by showing an example that deserves the ratings of 1, 2 and 6 respectively with explanations for the reasoning as shown in \Cref{fig:demo}.
We randomly selected 96 examples of ReplicaGrasp covering objects in all 48 receptacles in both upright and fallen orientations.
We showed each example to 5 different subjects and we had 30 unique subjects solve the task.
We filtered out cases which saw high inter-subject disagreement.

\cref{fig:mturk_viz} shows some examples of~\flex~generations evaluated by subjects. 
Participants generally found the results realistic -- for example, for \cref{fig:mturk_viz} b, a participant wrote: \textit{``The stretch of the hand inside the drawer is very realistic''}. 
In some cases where the subjects gave a low rating, for instance in \cref{fig:mturk_viz} c, we received interesting comments: \textit{``Doesn't need to squat to grab item''}. This reveals a shortcoming of our system wherein we do not measure the effort required to grasp an object. Explicitly modeling physical effort and its effect on the choice of the human's pose is an interesting direction that we leave as future work.

\section{Computational Budget}
\label{supp_comp_budget}
We performed speed and memory comparisons (averaged across 10 runs) for generating 16 different samples on a single RTX 2080 Ti GPU.
\flex~involves using pre-trained models simultaneously, the memory consumption is 3x (4.8 GB vs 1.4 GB).
\flex~takes around 8.5 minutes to generate 16 samples, while SAGA and GOAL take 6 and 1 minute respectively. We sacrifice computational budget for significantly better results.

\section{Diversity Analysis}
\label{supp_div_analysis}
\input{div}
\cref{tab:div} shows diversity metric computed for hand (no-full-body) and for full-body (no-hand) for all 3 methods.~\flex~has higher diversity in both, but the gains are significant for full-body.

\section{Choice of Optimization Framework}
\label{supp_optim}
\input{optim}
\flex~is agnostic to the choice of the optimization method. We used the recent Liu \textit{et al.}~\cite{liu2022landscape} which smooths the loss landscape for better convergence.
To validate this choice of a gradient-based optimization framework, we conduct experiments with non-gradient based methods described below:
\begin{itemize}[nosep,leftmargin=*]
    \item\textbf{Ranking:} Instead of optimization, we simply rank a large the batch of whole-body grasps produced by randomly sampling the optimization parameters.
    \item\textbf{CMA-ES:} Covariance matrix adaptation evolution strategy implemented from \href{https://pypi.org/project/cmaes/}{PyPI}.
\end{itemize}
Results are shown in~\cref{tab:optim}.
As expected,~\flex~is superior to both the baselines.

\section{Performance as a function of the number of optimization steps}
\label{supp_iter}
\begin{figure}[t!]
    \centering
    \includegraphics[width=0.8\linewidth]{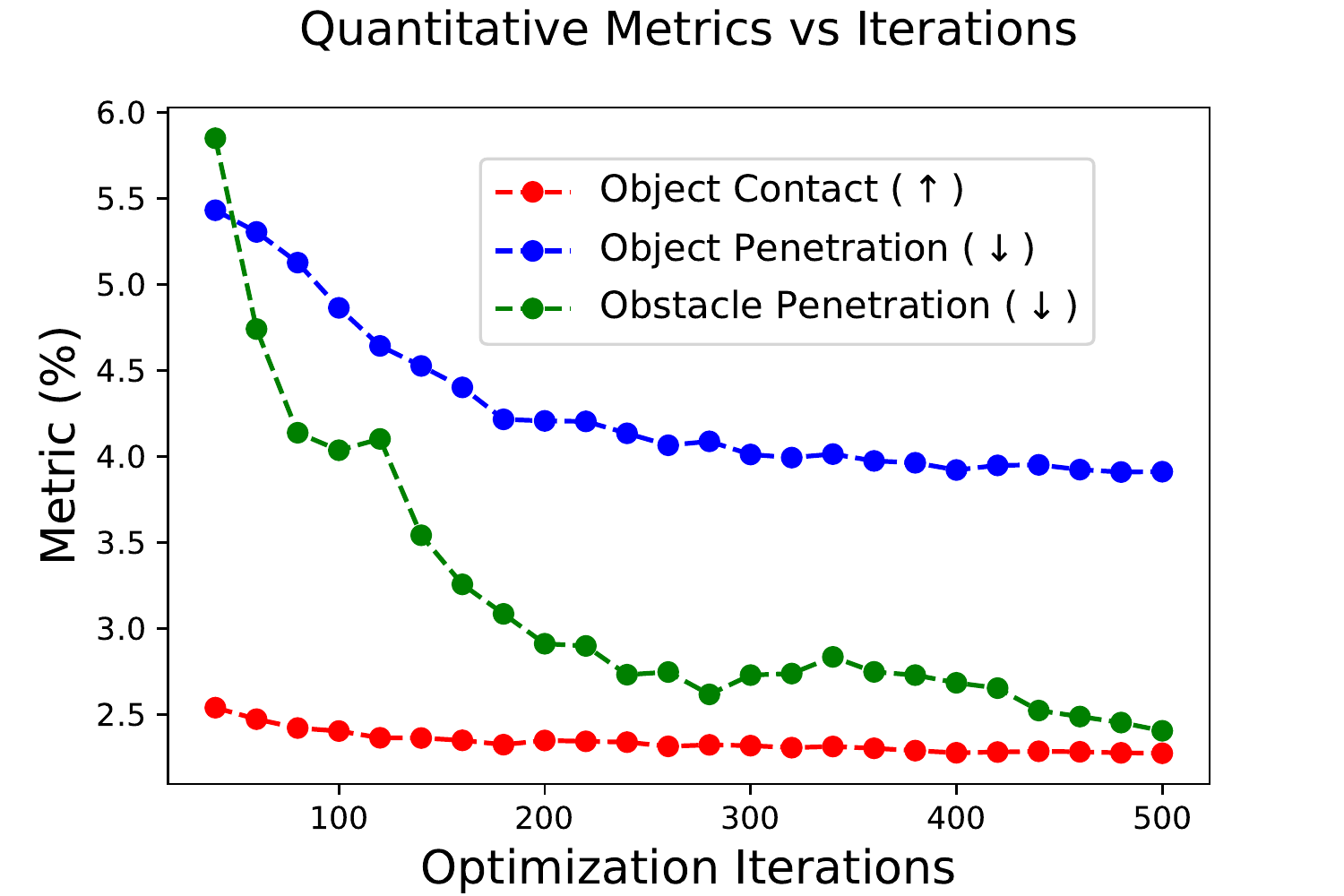}
    \caption{Object contact, penetration and Obstacle penetration metrics varying by iterations.}
    \label{fig:iter}
\end{figure}
\cref{fig:iter} shows average optimization metrics over different iterations. Object and obstacle penetration go down with training. Object contact stays largely unchanged.

%% file: div.tex
\begin{table}[t!]
\centering
\footnotesize
\tablestyle{2.5pt}{1.05}
\begin{tabularx}{1.0\linewidth}{lYYYY}
    \toprule
      & \multicolumn{2}{c}{Sample-wise $\uparrow$} & \multicolumn{2}{c}{Overall $\uparrow$} \\
      \cmidrule(lr){2-3}\cmidrule(lr){4-5}
      
     \multicolumn{1}{c}{Method} &
     Full Body & Right-hand &
     Full Body &
     Right-hand \\
    \midrule
    GOAL & 0.11 & 0.04 & 6.01 & 12.14  \\
    SAGA & 1.14 & 0.09 & 15.29 & 13.79 \\
    \flex~(ours) & \textbf{26.91} & \textbf{0.36} & \textbf{39.98} & \textbf{16.40} 
\end{tabularx}
\caption{Diversity analysis on GRAB (cm)}
\label{tab:div}
\end{table}

%% file: optim.tex
\begin{table}[t!]
\centering
\footnotesize
\tablestyle{2.5pt}{1.05}
\begin{tabularx}{\linewidth}{lYYYY}
    \toprule
     \multicolumn{1}{c}{\textbf{Method}} &
     \textbf{Obj Cont (\%) $\uparrow$} &
     \textbf{Obj Penet (\%) $\downarrow$} &
     \textbf{Obs Penet (\%) $\downarrow$} &
     \textbf{Ground (cm) $\downarrow$} \\
    \midrule
    Random Init & 0.15 & 35.06 & 2.02 & 60.32 \\
    CMA-ES & 0.03 & 17.39 & 2.03 & 58.25 \\
    \flex & \textbf{2.20} & \textbf{2.50} & \textbf{0.53} & \textbf{0.00} \\
    \bottomrule
\end{tabularx}
\caption{Comparison with Optimization methods.}
\label{tab:optim}
\end{table}